\documentclass[final,5p,times,twocolumn]{elsarticle}
\usepackage{amssymb}
\usepackage{amsmath}
\usepackage{subcaption}
\usepackage{soul}
\usepackage{xcolor}
\usepackage{ulem}
\usepackage{framed}
\usepackage{acronym}
\usepackage{multicol}
\usepackage{svg}
\usepackage[colorlinks]{hyperref}

\journal{Advanced Engineering Informatics}

\begin{document}

\begin{frontmatter}



\title{State-of-the-art review and synthesis: A requirement-based roadmap for standardized predictive maintenance automation using digital twin technologies}


\author[label1]{Sizhe Ma}
\author[label1]{Katherine A. Flanigan}
\author[label1]{Mario Berg\'es \fnref{label2}}

\affiliation[label1]{organization={Carnegie Mellon University},%Department and Organization
            addressline={5000 Forbes Avenu}, 
            city={Pittsburgh},
            postcode={15213}, 
            state={PA},
            country={USA}}
\fntext[label2]{Mario Berg\'es holds concurrent appointments at Carnegie Mellon University (CMU) and as an Amazon Scholar. This manuscript describes work at CMU and is not associated with Amazon.}

%% Abstract
\begin{abstract}
Recent digital advances have popularized predictive maintenance (PMx), offering enhanced efficiency, automation, accuracy, cost savings, and independence in maintenance processes. Yet, PMx continues to face numerous limitations such as poor explainability, sample inefficiency of data-driven methods, complexity of physics-based methods, and limited generalizability and scalability of knowledge-based methods. This paper proposes leveraging Digital Twins (DTs) to address these challenges and enable automated PMx adoption on a larger scale. While DTs have the potential to be transformative, they have not yet reached the maturity needed to bridge these gaps in a standardized manner. Without a standard definition guiding this evolution, the transformation lacks a solid foundation for development. This paper provides a requirement-based roadmap to support standardized PMx automation using DT technologies. Our systematic approach comprises two primary stages. First, we methodically identify the Informational Requirements (IRs) and Functional Requirements (FRs) for PMx, which serve as a foundation from which any unified framework must emerge. Our approach to defining and using IRs and FRs as the backbone of any PMx DT is supported by the proven success of these requirements as blueprints in other areas, such as product development in the software industry. Second, we conduct a thorough literature review across various fields to assess how these IRs and FRs are currently being applied within DTs, enabling us to identify specific areas where further research is needed to support the progress and maturation of requirement-based PMx DTs.
\end{abstract}

\begin{keyword}
Digital Twins \sep Predictive Maintenance \sep Technical Requirements \sep Artificial Intelligence



\end{keyword}

\end{frontmatter}
\begin{figure*}
\begin{framed}
\section*{List of Acronyms}
\begin{multicols}{2}
\begin{acronym}[AAAAAAAA]\itemsep0pt  % longest acronym to fix width

    \acro{ACM} {Association for Computing Machinery}
    \acro{AG} {Advisory Generation}
    \acro{AHU} {Air Handling Unit}
    \acro{AI} {Artificial Intelligence}
    \acro{CM} {Corrective Maintenance}
    \acro{DA} {Data Acquisition}
    \acro{DM} {Data Manipulation}
    \acro{DTh} Digital Thread
    \acro{DT} {Digital Twin}
    \acro{FDD} {Fault Detection and Diagnosis}
    \acro{FR} {Functional Requirement}
    \acro{HA} {Health Assessment}
    \acro{HVAC} {Heating, Ventilation, and Air Conditioning}
    \acro{HI} {Human Interface}
    \acro{IR} {Informational Requirement}
    \acro{ISO}{International Organization for Standardization}
    \acro{IoT} {Internet of Things}
    \acro{IEEE} {Institute of Electrical and Electronics Engineers}
    \acro{LIME} {Local Interpretable Model-agnostic Explanations}
    \acro{ML} {Machine Learning}
    \acro{OSA-CBM} {Open System Architecture for Condition-based Monitoring}
    \acro{PLM} {Product Lifecycle Management}
    \acro{PIML} {Physics-informed Machine Learning}
    \acro{PMx} {Predictive Maintenance}
    \acro{PvM} {Preventive Maintenance}
    \acro{PCA} {Principal Component Analysis}
    \acro{PA} {Prognostics Assessment}
    \acro{RUL} {Remaining Useful Life}
    \acro{SHAP} {Shapley Additive Explanations}
    \acro{SD} {State Detection}
    \acro{VAV} {Variable Air Volume}
\end{acronym}
\end{multicols}
\end{framed}
\end{figure*}

\section{Introduction}
\label{sec:Section1}

Effective maintenance is essential for ensuring the efficiency, safety, and longevity of a system and its assets. A number of
maintenance strategies have emerged to improve maintenance processes such as Predictive Maintenance (PMx), Corrective
Maintenance (CM), Preventive Maintenance (PvM), and Total Productive Maintenance~\cite{mobley2002introduction}. Of these
approaches, PMx is most suitable when the maintenance strategy requires the avoidance of run-to-failure circumstances. This is
the case for the vast majority of complex, real-world systems. PMx is a proactive maintenance strategy that utilizes data
analysis, artificial intelligence (AI), and other advanced technologies to predict when a system is likely to fail and take
necessary actions to prevent it. This approach uses real-time monitoring, data analysis, and historical data to predict
equipment failure, allowing maintenance teams to schedule repairs or replacements before breakdowns occur.

PMx has undergone significant advancements over the past decade, providing new ways to improve system
performance~\cite{SAKIB2018267}. These advancements have been made possible by the emergence and integration of new
technologies supporting the Internet of Things (IoT), greater volumes of data, and increases in computing
power~\cite{ran2019survey}. Increased data availability has resulted from the wide range of advanced technologies that are
increasingly being integrated into industrial processes and engineered systems. Referred to as the IoT, this comprises
networks of physical systems that are embedded with sensors, software, and connectivity, enabling them to collect and exchange
data with each other and with the Internet. PMx also relies on complex algorithms and models that can process large quantities
of data produced~\cite{SAKIB2018267,serradilla2022deep}. With increased computing resources, it is now possible to monitor,
analyze, and predict system behavior in near real time. Together, these advancements have led to more accurate predictions and
better-informed maintenance decision-making capabilities.

Despite these advancements, PMx faces a number of limitations inhibiting its development and widespread adoption. These
limitations are tied to the data-driven, physics-based, and knowledge-based modeling techniques and requirements used for PMx.
Here, we review a few of the limitations of each modeling approach. For data-driven PMx, failure data is scarce as systems are
typically not operated to failure, which adversely affects a model's failure prediction
capabilities~\cite{michele2020iot,wen2022recent}. Second, there is a lack of explainability in current models, which can raise
doubts about algorithmic accuracy and affect the trustworthiness of the decisions and recommendations
proposed~\cite{shukla2020opportunities}. Third, a large proportion of algorithms applied within PMx contexts are not sample
efficient and require vast amounts of data in order to achieve viable performance in practice~\cite{wen2022recent}. For
physics-based PMx, the complexity of models can be intractable. PMx relies on complex algorithms and models, which can be
time-consuming or even unfeasible to develop~\cite{NUNES202353}. This is compounded by the complexity of near real-time
monitoring, analysis, and prediction requirements, as well as the high level of expertise needed to implement such
requirements~\cite{aivaliotis2019methodology}. For knowledge-based modeling techniques, existing models often have limited
generalizability and scalability because they are typically developed for specific types of equipment with manual
efforts~\cite{fink2020pmx}. To help combat these limitations, hybrid approaches such as Physics-informed Machine Learning
(PIML), scientific machine learning, and hybrid models have emerged~\cite{sepe2021physics,nascimento2019fleet}.
Although these approaches make acquiring certain system states and properties easier, they typically make use of or inform
only a limited part of the physical knowledge. When using PIML, for example, knowledge or rules act as implicit boundaries
that are embedded into models such as artificial neural networks, which remain incapable of addressing the
aforementioned limitations. 

The PMx field -- like numerous other fields across engineering -- is contemplating the forthcoming digital transformation as a
potential avenue to mitigate these challenges. Specifically, the advent of Digital Twins (DTs), which are dynamic digital
representations of physical assets, presents significant opportunities to address these obstacles and create a seamlessly integrated
intelligent system that can predict and prevent equipment failures, optimize maintenance tasks, and vastly improve the
efficiency and reliability of the entire operational process~\cite{NAP26894}. Already, we have seen practical applications of
DTs making a difference across sectors of PMx. In the aerospace industry, DTs have been employed to simulate the behavior and
performance of critical components on aircraft under diverse operational conditions by incorporating both data-driven and
physics-based models. This has enabled preemptive detection of potential failures and optimization of maintenance
scheduling~\cite{zaccaria2018twin,li2017dynamic}. In manufacturing, DTs are being used not only to predict equipment failures
but also to provide real-time insights into the manufacturing process, enhance product quality, and streamline workflow.
Applications have been detailed in different resolutions, including advanced machinery, assembly lines, and shop floors, which
explore the potential of DTs in high-precision manufacturing environments to inform decisions and reduce downtime and
operational costs \mbox{\cite{tao2017twin,lv2023twintwin,met11050708}.}  

However, the literature often reflects fragmented adoption of DTs, primarily focusing on specific predictive tasks within
limited sectors such as aerospace and manufacturing. This underscores a significant gap in the comprehensive study
of DTs' broader applicability and integration into PMx frameworks across various industries~\cite{van2022predictive}. Despite
the acknowledged potential of DTs to enhance predictive analytics and maintenance scheduling, existing studies rarely
explore how these technologies can be seamlessly integrated into the diverse operational contexts typical of different
industrial sectors~\cite{liu2018digitaltwin,falekas2021twin}. This lack of extensive research highlights the need for more
detailed studies that not only demonstrate the technical capabilities of DTs but also their practical effectiveness and
scalability~\cite{errandonea2020twin}. Addressing this need will provide clearer insights into how DTs can be tailored and
effectively implemented to meet the unique challenges of various industries, thereby optimizing maintenance processes and
enhancing system reliability across the board.

Compounding these challenges is the lack of a consistent definition of what constitutes a DT. Despite the nearly exponential rise in DT literature published over the past decade~\cite{kunzer2022digital}, the definition
of what a DT is and consists of is constantly changing, often times to stake novelty claims to the idea in specific technical
fields. This lack of consensus in defining DTs -- both across fields and within specific disciplines -- has led to a diverse
interpretation of the concept, thereby complicating the realization of the envisioned benefits and its ability to readily
overcome the challenges facing PMx automation. This discrepancy leads to inefficiencies in knowledge transfer, collaboration,
and progress across the field. A primary challenge impeding the widespread adoption of PMx DTs across the industrial sector
lies in the variability inherent in the requirements for PMx systems as perceived by different
stakeholders~\cite{li2020toward}. 

\begin{figure*}
\centering
\includegraphics[scale=0.1]{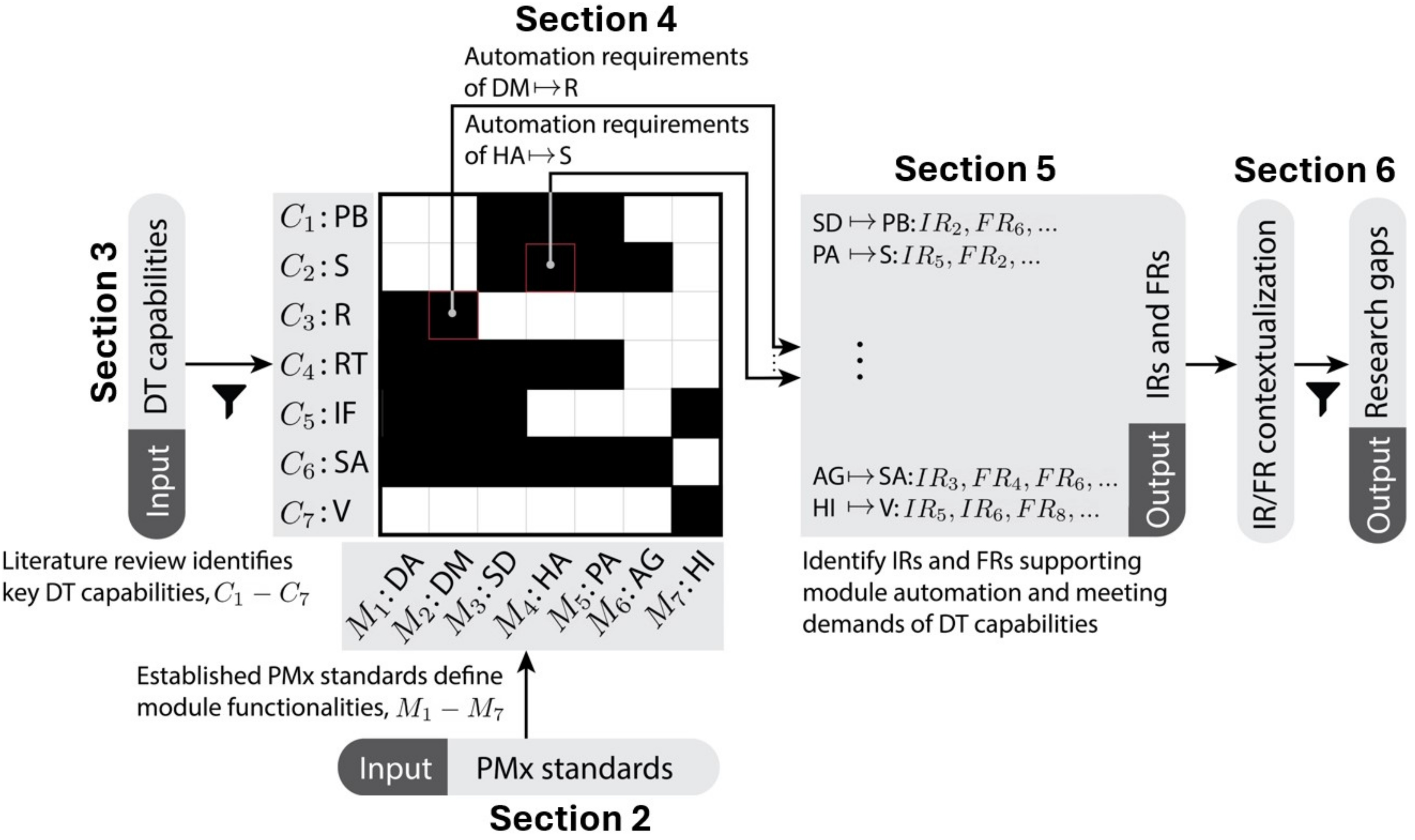}
\caption{Paper overview including key methodological steps. Module acronyms are provided in Section~\ref{sec:Section2}, and DT capability acronyms are provided in Section~\ref{sec:Section3}.}
\label{fig:Figure1}
\end{figure*}

The establishment of more precise and standardized definitions and requirements supporting PMx automation using DTs is
instrumental in facilitating a clear understanding among researchers, industry practitioners, and engineers of the exact
demands that DTs necessitate. In the maintenance field, standardized requirements have always played a pivotal role in
promoting efficiency and ensuring the reliability of systems~\cite{li2020toward,nordal2021model}. Surprisingly, Informational
Requirements (IRs) and Functional Requirements (FRs), fundamental to such processes, have never been formally presented as
explicit requirements in this context previously, signifying a missed opportunity to formalize PMx DTs. The development of
clear IRs and FRs would aid the formulation of a more consistent and widely applicable definition of PMx DTs. Looking across
other fields that have faced similar challenges, the system and software industry have successfully developed and used IRs and
FRs to frame and guide the development of systems and products~\cite{lana2021data,sommerville2009deriving}; these IRs and FRs
have provided the basis for coherent, agreed-upon criteria for developments supporting autonomy that minimize barriers to
interdisciplinary collaboration. Introducing well-defined IRs and FRs for PMx would clarify informational needs and behavioral responses. This would enable the design of a robust and adaptable pipeline, capable of effectively harnessing physics-based, data-driven, or hybrid approaches. By addressing key challenges like enhancing model explainability, this approach would foster a more reliable and trustworthy data-driven strategy~\cite{lana2021data}. IRs and FRs offer a structured way to navigate the complexities of PMx and bridge
the gap between theory and application, making models more understandable, trustworthy, and versatile. This clarity of
requirements would further facilitate the integration of hybrid methodologies, fostering more effective utilization of
physical knowledge and enabling the development of more sophisticated and reliable PMx systems. We propose that, within the
context of PMx, such a classification and standardization would not only benefit the academic world by establishing clear
definitions for what has been, to date, largely open to interpretation, but also benefit industry by overcoming barriers to
scaling and extending the implementation of PMx in practice. 

In this paper, we methodically identify and explicitly define the IRs and FRs for PMx systems, which serve as a roadmap from which any unified PMx DT framework must emerge. The proposed methodology recognizes the IRs and FRs as the mapping between PMx standards and DT capabilities. These, in turn, support the automation of PMx within a DT framework.  In other words, existing standards and DT capabilities are considered as boundary conditions that constrain the selection of the IRs and FRs. This effort differentiates our work by not only synthesizing the current state of the art but also by critically analyzing the gaps and integration challenges between PMx strategies and DT technologies. Unlike work that focuses solely on technological advancements, our approach delves into the practical implications of these technologies within operational contexts and their alignment with long-term maintenance strategies. This dual focus allows us to propose a framework that considers both the theoretical and applied aspects of DTs in the PMx domain, setting a new benchmark for holistic requirement-based standardization in the field. 

In Section~\ref{sec:Section2}, we first provide an overview of the existing standards that establish general guidelines for PMx. These standards define seven modules within condition monitoring processes, along with the general inputs and outputs of those seven modules. In Section~\ref{sec:Section3}, we then identify a distinct set of DT capabilities unique to PMx. Here, we detail our screening method used to select DT capabilities that are critical for PMx, employing a rigorous review process that ensures only the most relevant capabilities are included. Each DT capability necessitates specific module functionality. We scope the demands of the DT capability and identify the IRs and FRs that support scalable and automated operation of the module to meet these requirements. This procedure is outlined in detail in Section~\ref{sec:Section4}. Once the IRs and FRs have been exhaustively identified, we draw upon industry standards, state-of-the-art PMx literature, and expert opinions to complement these requirements with concrete examples, contextualizing each requirement. Subsequently, a similar screening method is also applied in Section~\ref{sec:Section5} when we examine DT-enabled PMx, ensuring a consistent and thorough review process throughout the paper. This helps in determining how the IRs and FRs are currently being utilized within PMx DTs, enabling the identification, scoping, and prioritization of the key gaps and research areas needed to support an automated PMx DT framework. The paper concludes with summarizing recommendations and next steps. An overview of the methodology proposed in this paper is depicted in Fig.~\ref{fig:Figure1}. 

Our primary contributions in this paper include:
\begin{itemize}
    \item We identify a distinct set of DT capabilities that are uniquely associated with PMx, addressing the gap in previous studies that generalize DT capabilities without specific tailoring for PMx applications.
    \item We identify and define IRs and FRs specifically tailored for PMx systems, a significant advancement over existing literature that lacks such formal presentation and explicit definitions. This delineation is justified with real-world illustrations from industry benchmarks, cutting-edge PMx research, and insights from domain experts.
    \item We develop an innovative mapping methodology that bridges PMx standards with DT capabilities, facilitated by the identified IRs and FRs. This methodology fills the previous gap of lacking a systematic approach to cohesively develop DT-enabled PMx solutions, facilitating more integrated and effective implementation.
    \item We conduct a comprehensive literature review to assess how IRs and FRs are currently utilized within PMx DTs, identifying and addressing critical gaps that previous research has overlooked, particularly regarding the practical implications and effectiveness of these requirements in DT environments.
\end{itemize}

%%%%%%%%%%%%%%%%%%%%%
%%%%% SECTION 2 %%%%%
%%%%%%%%%%%%%%%%%%%%%

\section{Predictive Maintenance and Related Strategies}
\label{sec:Section2}

Maintenance is essential for ensuring the safe and economically efficient long-term use of equipment across various industries. Significant progress has been made in improving maintenance processes over the past two decades. As previously mentioned, three primary maintenance strategies are used to trigger maintenance actions: CM, PvM, and PMx. 

\begin{figure}
\centering
\begin{subfigure}[b]{0.23\textwidth}
   \includegraphics[width=1\linewidth]{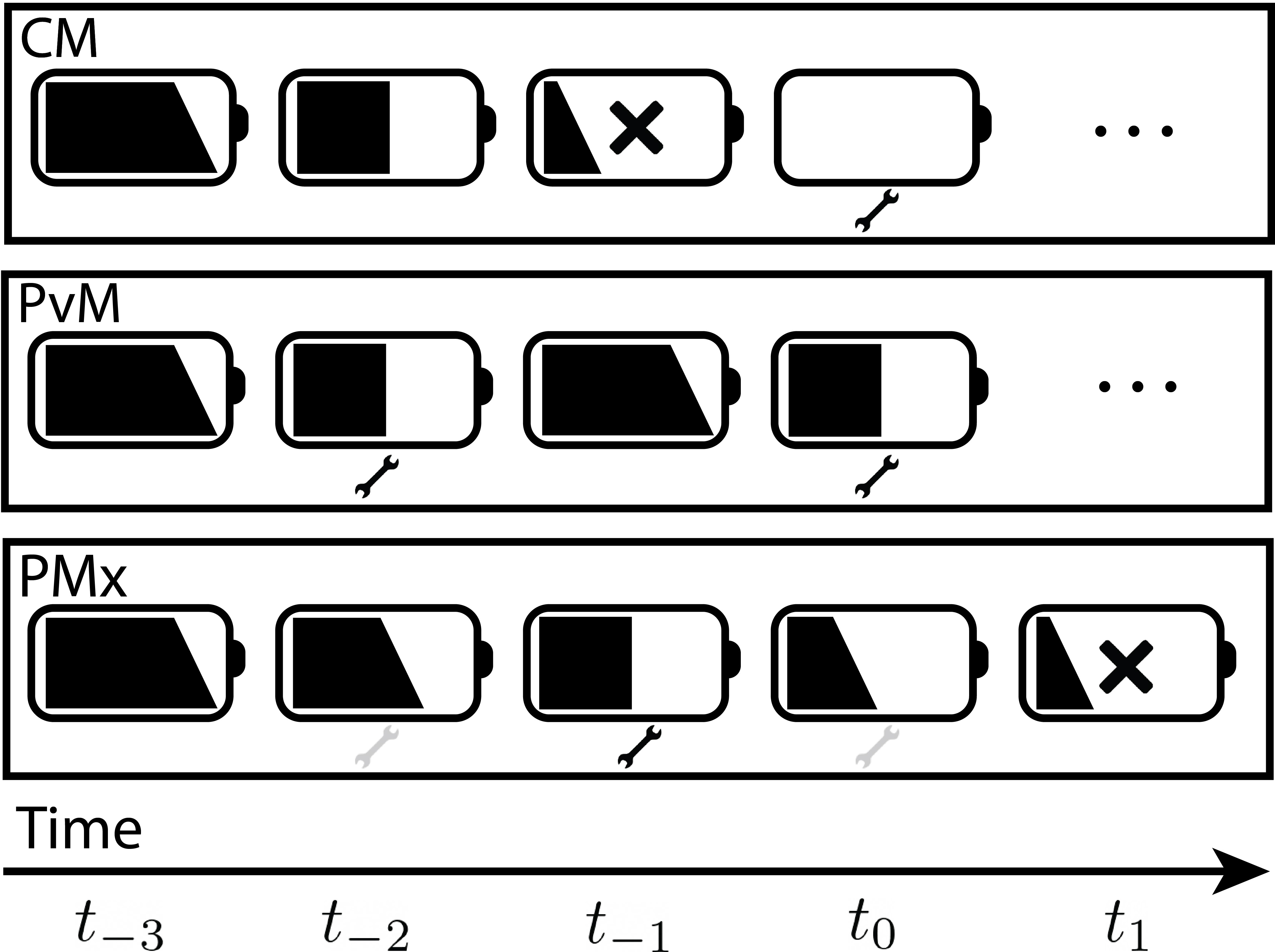}
   \caption{}
   \label{fig:Figure2a} 
\end{subfigure}
\hfill
\begin{subfigure}[b]{0.23\textwidth}
   \includegraphics[width=1\linewidth]{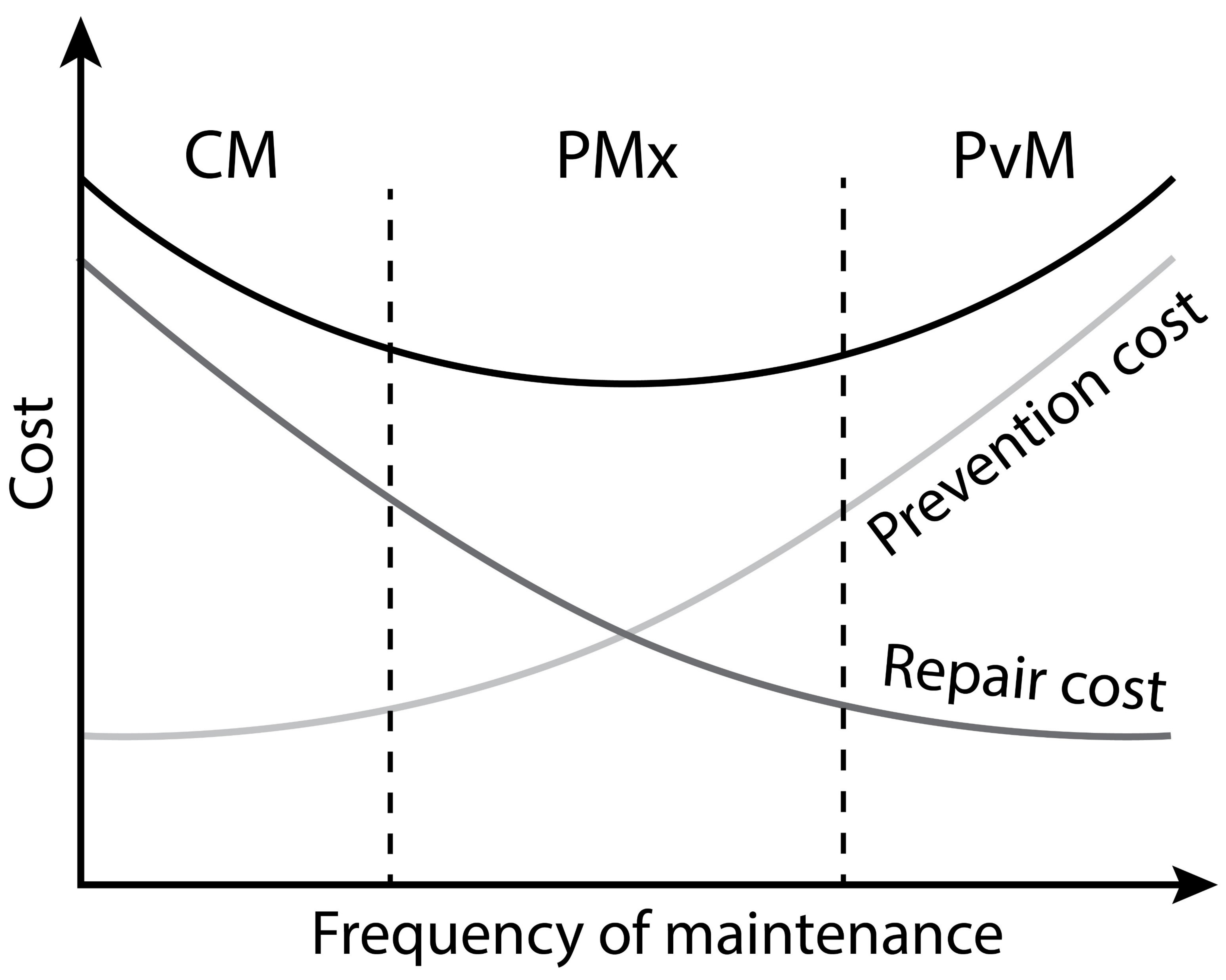}
   \caption{}
   \label{fig:Figure2b}
\end{subfigure}

\caption{A comparison between different maintenance techniques with respect to (a) the underlying mechanisms and (b) the economy \cite{ran2019survey}.}
\end{figure}

As illustrated in Fig.~\ref{fig:Figure2a}, CM triggers actions once component or system faults occur, PvM schedules maintenance activities on a regular basis to lessen the likelihood of failures, and PMx aims to predict the failure of equipment based on the estimation of future states in order to propose maintenance recommendations out of a range of options with respect to resource availability and economic factors. Fig.~\ref{fig:Figure2b} compares these three strategies, illustrating the costs incurred to address repairs as well as those associated with preventive activities. Fig.~\ref{fig:Figure2b} suggests that CM suffers from excessive replacement costs and collateral damage, while PvM may lead to unnecessary repairs and wasted resources. 

PMx, on the other hand, balances these strategies by determining an optimal time to trigger maintenance
actions~\cite{jimenez2019system}. Deviating from this schedule could result in either wasted resources or an increased
likelihood of failures. Conventionally, PMx draws on information about the system's condition (e.g.,~near real-time operating
conditions, configurations) to optimize the maintenance schedule. This is typically done in a predictive manner by identifying
future usage patterns and Remaining Useful Life (RUL). The proliferation of Machine Learning (ML) algorithms -- which have
emerged in recent years in response to increases in computational power -- have led to significant advancements in AI-guided
optimization for PMx and play an integral role in shaping the IRs and FRs.

\begin{figure}
\centering
\includegraphics[width=1\linewidth]{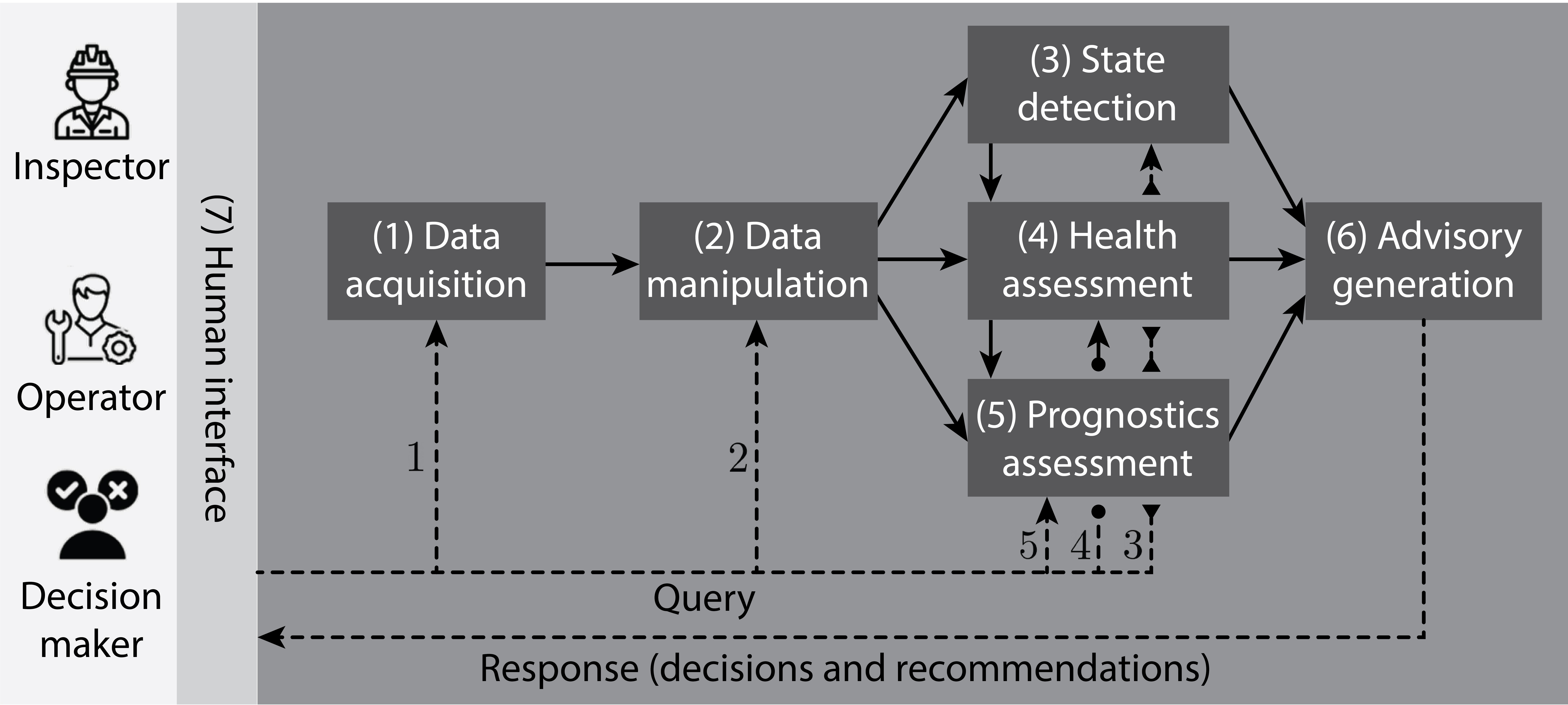}
\caption{PMx module layout and key interdependencies adopted from OSA-CBM.}
\label{fig:Figure3}
\end{figure}

The International Organization for Standardization (ISO) has issued several standards related to PMx, most notably ISO-13374~\cite{ISO13374}, which provides general guidelines for software specifications concerning data processing,
communication, and presentation of machine condition monitoring and diagnostic information. ISO-13374 defines six
functionalities within condition monitoring processes, as well as general inputs and outputs of those six functionalities.
These functionalities encompass the four primary levels of advanced maintenance strategies, as previously defined~\cite{haarman2017predictive}, each characterized by distinct methodologies: Level 1 involves visual inspection, Level 2 employs instrument inspection, Level 3 utilizes near real-time condition monitoring, and Level 4 integrates continuous monitoring with predictive
methods. The Machinery Information Management Open Systems Alliance's Open System Architecture for Condition-based Monitoring (OSA-CBM)~\cite{OSACBM}
further extends ISO-13374 by adding data structures and defining the ``Human Interface'' as a functionality, increasing the
practical application capabilities of the original six functionalities defined by the ISO standard. These standards provide a
standardized and operational foundation from which the IRs and FRs will be rooted. An illustrative architecture supported by the
OSA-CBM is depicted in Fig.~\ref{fig:Figure3}, where the functionalities are defined as follows:

\begin{enumerate}
    \item Data Acquisition (DA): Conversion of the system's physical conditions into digital form (e.g.,~data) and collection according to specific IoT protocols.
    \item Data Manipulation (DM): Pre-processing the collected data (e.g.,~signals) into forms suitable for future uses. These processes can vary, and can include denoising and feature extraction, for example.
  \item State Detection (SD): Comparing processed features or raw data against reference values (e.g.,~run-to-failure data, threshold values) to determine if the assets are functioning as intended, often generating binary
indicators~\cite{kamat2020anomaly,li2023twin}. Alarms can be triggered if necessary, such as in an anomaly detection system
that flags a machine's temperature when it exceeds a threshold.
  \item Health Assessment (HA): Evaluating the system's health using standardized or customized indicators, typically requiring data like usage history and current operational status~\cite{ran2019survey,hosamo2022digital}.
For example, a diagnostic system that assesses a machine's current health level by comparing real-time and historical data.
  \item Prognostics Assessment (PA): Projecting the system's future health (e.g.,~future health level, RUL) by considering usage profiles, maintenance team schedules, and future resource
availability~\cite{wen2022recent,zhang2023rul}. For instance, a prognostic system that predicts a machine's RUL by forecasting
future health levels.
    \item Advisory Generation (AG): Providing decision makers (e.g.,~mechanics, assets managers) with a list of maintenance decisions and recommendations to choose from. This module can operate through manual inputs, where human operators analyze recommendations, or through automated feedback control that dynamically implements decisions based on real-time data.
    \item Human Interface (HI): Supplying an information system that allows operators to input data or receive outputs from previous modules, facilitating interaction with the system components.
\end{enumerate}

The modules outlined above are not unique to PMx but are broadly applicable across various fields. For example, SD in
industrial applications is comparable to quality control in manufacturing, where both involve monitoring parameters to
ensure they meet predefined standards~\cite{stojanovic2016data}. Similarly, HA is analogous to system diagnostics in
automotive engineering, which focuses on evaluating the operational status of vehicle
systems~\cite{JEGADEESHWARAN2015436}. Additionally, PA is similar to forecasting in multiple engineering fields,
where predictive models estimate future conditions based on historical data~\cite{berghout2020aircraft}. These
parallels highlight the versatility and adaptability of these modules, making them integral to various technological and
industrial domains.

Among these seven modules, SD, HA, and PA constitute the `three pillars' of PMx, corresponding to anomaly detection, diagnosis, and prognosis, respectively. A maintenance task must include at least one of these three modules to be considered a PMx task. We present this standard because it is well-established and satisfies two key characteristics that support IR and FR identification in Section~\ref{sec:Section4}: (1) comprehensiveness: the standard provides detailed steps for most PMx tasks without relying on new or proprietary architectures and (2) explicit input/output: the standard provides clear information flow between modules.

Within these three `pillars,' while SD and DA are equally foundational to the operational workflow of PMx, it is PA that
truly defines the predictive nature of this strategy. Unlike reactive or basic condition-based maintenance approaches, PA
extends maintenance capabilities by forecasting the future health status of assets. This forward-looking
approach, central to PMx, leverages advanced predictive analytics to estimate critical metrics representing future health
levels for maintenance planning~\cite{lei2018prognosis}. For example, focusing on RUL estimation demonstrates the sophistication and
evolution of these predictive techniques: once reliant on deterministic models, RUL estimation has advanced
significantly~\cite{si2011data,ferreira2022remaining}. Modern data-driven methods, such as deep learning, now enhance accuracy but
also require larger datasets~\cite{wang2020deeplearning}, leading to a recent shift towards hybrid models that combine data
insights with physical principles to improve both interpretability and operational efficiency~\cite{li2024informed}.

%%%%%%%%%%%%%%%%%%%%%
%%%%% SECTION 3 %%%%%
%%%%%%%%%%%%%%%%%%%%%

\section{Digital Twin and Digital Thread Capabilities}
\label{sec:Section3}

\begin{figure*}
\centering
\includegraphics[width=0.8\linewidth]{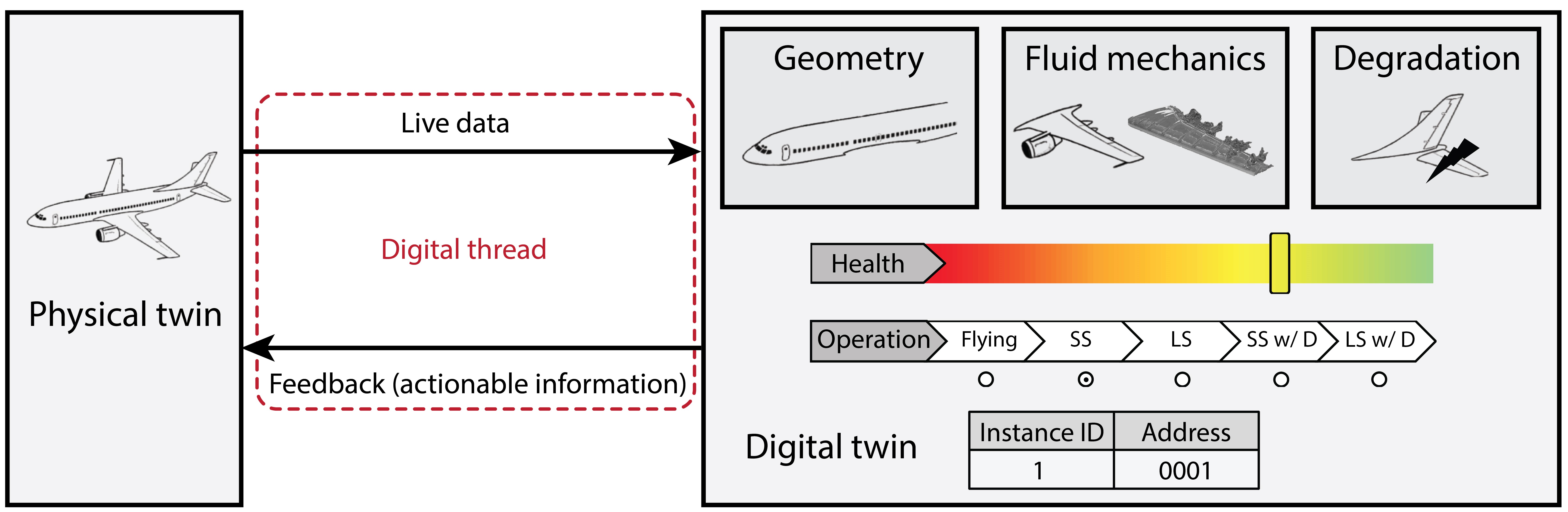}
\caption{High-level overview of a DT supporting aviation-related PMx with key components highlighted. We include the following notation, SS: short stay, LS: long stay, SS w/ D: short stay with disturbance, LS w/ D: long stay with disturbance.}
\label{fig:Figure4}
\end{figure*}

In 2002, Dr. Micheal Grieves presented the `Conceptual Ideal for Product Lifecycle Management (PLM),' a seminal work on the
evolution of DTs that was subsequently formalized and published in 2005~\cite{grieves2005product}. Within the PLM framework,
three key elements -- physical space, virtual space, and the information flow between them -- are defined as
fundamental components that all DTs must contain. Fig.~\ref{fig:Figure4} illustrates how these elements operate and interact
within the context of PMx. This figure also introduces the concept of the Digital Thread (DTh), which, in the context of PMx, refers to
the interconnected flow -- either wired or wireless -- and systematic integration of data across an asset or system's life
cycle~\cite{kunzer2022digital}. This integration includes data generated during the design, operation, and maintenance phases,
providing a comprehensive understanding of the asset's health and performance. Conceptualizing DThs involves envisioning them
as threads that weave together information such as sensor data, operational data, maintenance records, and inspection
data, forming a dynamic model that evolves throughout the asset or system's lifecycle.

Following the initial conceptualization of DTs, there were few organized definitions, with different fields adapting the
term to fit their specific needs. It was not until the recent surge in DT popularity -- driven
by advancements in IoT, AI, and ML, along with increased data availability -- that Grieves proposed a more succinct DT
definition as follows: \textit{`a set of virtual information constructs that fully describes a potential or actual physical
manufactured product from the micro atomic level to the macro geometrical level'}~\cite{grieves2017digital}. This definition
was supported by a structured methodology now known as the DT framework, which has been instrumental in
the creation, implementation, and management of DTs throughout their lifecycle. 

The DT framework involves collecting in-use information to inform the creation phase of the twinned asset and running near real-time simulations to help mitigate unpredictable and undesirable behaviors. These two capabilities, known as `design in use' and `what-if' simulation, have evolved into widely applied features of DTs and DT frameworks in both academia and industry. 

The rationale for emphasizing the realization of these two capabilities goes beyond highlighting the growing
significance and application of DT-related technology. It also underscores the diverse development of DTs across various fields,
which has led to challenges in establishing a universally accepted, consistent definition of DTs. As Grieves'
two visions have materialized, different industries have adopted them in unique ways to meet their specific needs. For
example, the aviation and automotive industries, which each deal with complex systems and stringent safety requirements, have
primarily focused on leveraging the `what-if' simulation capabilities of DTs~\cite{barricelli2019twin}. In contrast, the
manufacturing industry has emphasized the `design-in-use' DT capability, capturing real-time operational data to inform and improve manufacturing processes and product design~\cite{HONGLIM202289}. Meanwhile,
the medical and healthcare fields have found a balance in both aspects of DTs~\cite{HALEEM202328}; `what-if' simulations support
surgical planning and medical training, while the `design-in-use' approach optimizes healthcare system operations. 

These two capabilities serve as illustrative examples as the scope of DTs continues to expand, incorporating additional features such as visualization, interoperability, and advanced human-system interaction, among others. As DTs evolve, industries are increasingly embracing and refining specific capabilities tailored to their unique needs and objectives, rather than adopting a consistent and unified framework.

Even within a single domain such as PMx, DT capabilities are often shaped by immediate, application-specific needs. This
approach prioritizes addressing current challenges faced by stakeholders, sometimes at the expense of long-term strategic considerations
and future system evolution. For instance, PMx tasks focused on prognosis require the ability to predict potential future
states or RUL~\cite{venkatesan2019health,karve2020digital}, emphasizing `what-if' simulation capabilities, in contrast to
PMx tasks centered on outlier detection. In the automobile industry, there is a need for multi-physics modeling to facilitate
collaboration among various systems like electronics, electromagnetics, and brake pads~\cite{magargle2017simulation}. This
sharply contrasts with bridge maintenance, which prioritizes long-term endurance and structural stability, relying
predominantly on the physics of material stress and strain~\cite{ye2019twin}. While the focus of DTs varies substantially across
PMx applications, a distinct set of capabilities unique to PMx can be identified.  These capabilities are distilled from a
comprehensive literature review of DT capabilities across PMx application areas. The selected literature is based on
stringent criteria ensuring relevance and comprehensiveness, with the filtering process involving three steps: string
search, abstract filtering, and content filtering. At each step, search criteria are refined to eliminate irrelevant
articles and proceedings.

\noindent \textit{Stage 1: String Search}\\
We first define relevant strings (i.e., the input search query) and databases to constrain the review scale. The search begins by utilizing Google Scholar\footnote{Google Scholar [website], scholar.google.com, (last accessed September 2023).} as it provides both a high volume and free resources to evaluate and update the string. After finalizing the query, the search is performed on four different databases: Springer\footnote{Springer Digital Library [website], www.springer.com/gp, (last accessed September 2023).}, ScienceDirect\footnote{ScienceDirect Digital Library [website], www.sciencedirect.com, (last accessed September 2023).}, Association for Computing Machinery (ACM) Library\footnote{ACM Digital Library [website], dl.acm.org, (last accessed September 2023).}, and IEEE Xplore\footnote{IEEE Xplore Digital Library [website], ieeexplore.ieee.org/Xplore, (last accessed September 2023).}. Studies that---while not being published on these four platforms---have garnered more than 50 citations according to Google Scholar, are included and categorized as ``others.'' The area of interest is then broadened to capture important works that do not explicitly incorporate these terms, but that use commonly associated or synonymous words.  The final search query is as follows:

\begin{quote}
(``digital twin'' OR ``cyber-physical system'') AND (``review'' OR ``survey'' OR ``state of the art'')
\end{quote}

\begin{table}[h]
\caption{Selection criteria for filtering abstracts for the DT capabilities review on PMx.}
\setlength{\tabcolsep}{3pt}
\centering
\begin{tabular}{|p{40pt}|p{195pt}|}
\hline
Identifier&
Criterion\\
\hline
A-CR1&
The study must be a comprehensive review or survey article published after 2012 in English\\
A-CR2&
The study must explicitly focus on the capabilities, applications, or functions of DTs\\
A-CR3&
The study must present a clear aim and, if any, results or conclusions drawn from the review of the DT literature\\
A-CR4&
The study must not be a book, technical report, dissertation, or thesis\\
A-CR5&
The study must discuss DTs in the context of one or more real-world applications or case studies\\
\hline
\end{tabular}
\label{Table1}
\end{table}

\begin{table}[h]
\caption{Selection criteria for filtering content for the DT capabilities review on PMx.}
\setlength{\tabcolsep}{3pt}
\centering
\begin{tabular}{|p{40pt}|p{195pt}|}
\hline
Identifier&
Criterion\\
\hline
    C-CR1 & The review must offer a comprehensive review of DT-related terminologies and concepts\\
    C-CR2 & The review must provide a clear objective for the review (e.g., taxonomy, comparison, state-of-the-art analysis)\\
    C-CR3 & The review must present and discuss findings in relation to the capabilities of the DT\\
    C-CR4 & The review must highlight current challenges and future directions in DT capabilities\\
    C-CR5 & The review must provide an overview of the application contexts and sectors in which the DT capabilities are applied\\
    C-CR6 & The review must cite reputable sources or studies to substantiate its findings on DT capabilities\\
    C-CR7 & The review must be highly recognized, as indicated by a substantial number of citations $(\geq 500)$, signaling its importance and influence in the DT field\\
\hline
\end{tabular}
\label{Table2}
\end{table}

\noindent \textit{Stage 2: Abstract Filtering}\\
Next, we perform filtering based on each paper's abstract. Several abstract criteria (A-CR) are applied to exclude papers outside of the search scope. These criteria are provided in Table \ref{Table1}.\\

\noindent \textit{Stage 3: Content Filtering}\\ 
Following the same strategies as the abstract filtering, the content criteria (C-CR) outlined in Table \ref{Table2} are applied during the reading of each study's content. This stage guarantees that the manuscripts fall within the relevant scope.\\

Based on this literature review, we present a list of the most frequently utilized DT capabilities within the PMx field \cite{barricelli2019twin,fuller2020twin,JONES202036,KRITZINGER20181016,LIU2021346,qi2018digital}:

\begin{enumerate}
  \item Physics-based (PB): A DT is considered physics-based if it provides a high-fidelity representation of the physical processes intrinsic to the asset~\cite{LIU2021346}. For example, finite element analysis is commonly employed to precisely simulate stress or temperature distributions in structural, fluid, and thermal mechanics. Physics-based capabilities support state detection, diagnosis, and prognosis tasks, enhancing the precision of failure mode predictions, especially when failure data is limited.
  \item Simulation (S): A DT supports simulation capabilities if it can reproduce real-world system behaviors under diverse conditions~\cite{barricelli2019twin}. The extent and focus of the simulations may vary depending on the task. For instance, a diagnosis task may require simulating the current time step, whereas a prognosis task might involve simulating future steps. Common across all implementations is the ability to evaluate asset status and the impact of various maintenance strategies without implementing these strategies on the actual physical system. This capability is particularly valuable when certain scenarios are impractical or costly to replicate in the real world.
  \item Replication (R): Replication, also known as ``representation,'' refers to a DT's ability to support a digital model that accurately mimics the physical properties and design details of its counterpart~\cite{LIU2021346}. The DT should include geometries, materials, and structural properties, encompassing both major and minor system components. Unlike physics-based and simulation capabilities, replication does not incorporate dynamic system behavior or physics laws. Instead, it provides a static, highly detailed model. High-fidelity representations can enhance the accuracy of system behavior predictions for state detection, diagnosis, and prognosis, improving the modeling of component interactions that influence overall system behavior.
  \item Real-time monitoring (RT): Real-time monitoring refers to a DT's ability to track the status of a physical system in real time, providing up-to-date insights into its condition~\cite{qi2018digital}. Real-time monitoring supports the prompt detection of response changes that could signify the onset of a failure, enabling maintenance to be scheduled before failure occurs.
  \item Integrated information (IF): A DT supports integrated information if it combines and synchronizes data from multiple sources (e.g., sensors, maintenance records, run-to-failure data), even if that data belongs to multiple stakeholders~\cite{barricelli2019twin}. Integrated information capabilities are essential for maintaining a complete, accurate, and current understanding of system states. The integration of diverse data not only enriches the system's knowledge base but also offers crucial insights for subsequent operations, known as downstream tasks. These downstream tasks can include analytics, predictive modeling, decision-making processes, and the execution of maintenance actions, all of which rely on the comprehensive and coordinated data provided by integrated information. Thus, integrated information is vital for ensuring seamless information flow and enhancing the effectiveness and efficiency of subsequent operations in the PMx pipeline.

  \item System automation (SA): A DT possesses system autonomy if it can operate and make decisions independently based on preset rules or ML algorithms~\cite{barricelli2019twin}. The degree to which an autonomous system supports and automates PMx functionalities determines its level of autonomy~\cite{flanigan2022autonomy}. A DT's system autonomy enables more efficient, responsive, and effective PMx processes, reducing reliance on manual monitoring and interventions and allowing for faster, more proactive maintenance actions.
  \item Visualization (V): Visualization refers to a DT's ability to represent the system's status and states in an intuitive and easily interpretable manner. Visualization capabilities often leverage graphical and user-friendly interfaces, such as digital dashboards~\cite{KRITZINGER20181016}. The level of detail conveyed within the visualization largely depends on the specific demands of the stakeholders, and the representation is influenced by the nature of the task at hand. For instance, state detection, diagnosis, and prognosis tasks each have unique quantities of interest and required levels of detail, which significantly impact the presentation of visualization.
\end{enumerate}

%%%%%%%%%%%%%%%%%%%%%
%%%%% SECTION 4 %%%%%
%%%%%%%%%%%%%%%%%%%%%

\section{Requirements Supporting Predictive Maintenance Autonomy} 
\label{sec:Section4}

%%%%%%%%%%%%%%%%%%%%%%%
%%%%% SECTION 4 A %%%%%
%%%%%%%%%%%%%%%%%%%%%%%

\subsection{Procedure for Identifying Requirements}
\label{sec:Section4-A}
This section provides an overview of the systematic procedure employed to identify the IRs and FRs that support PMx autonomy standardization. We define the IRs and FRs as the mapping between PMx standards identified in Section~\ref{sec:Section2} and DT capabilities identified in Section~\ref{sec:Section3}, with the IRs and FRs facilitating the automation of PMx within a DT framework (Fig.~\ref{fig:Figure1}). For each DT capability that requires a module functionality (e.g., DA, DM, SD), we determine the demands of the DT capability and identify the IRs and FRs necessary to support the scalable and automated operation of the module. Section~\ref{sec:Section4-A-1} first highlights the intersection between PMx modules and DT capabilities, providing a comprehensive `capability' grid that maps the requirements of each module to the unique capabilities offered by DTs. Section~\ref{sec:Section4-A-2} then outlines the procedure for identifying IRs and IRs within the capability grid. Section~\ref{sec:Section4-A-3} addresses the variability in asset diversity across different industries and its impact on implementing the PMx pipeline, placing the diverse range of assets on a spectrum of uniformity and explaning the examples used to contextualize the IRs and FRs defined in Section~\ref{sec:Section4-B} and Section~\ref{sec:Section4-C}.

\subsubsection{Capability Grid Definition} 
\label{sec:Section4-A-1}
\begin{figure}
\centering
\includegraphics[scale=0.9]{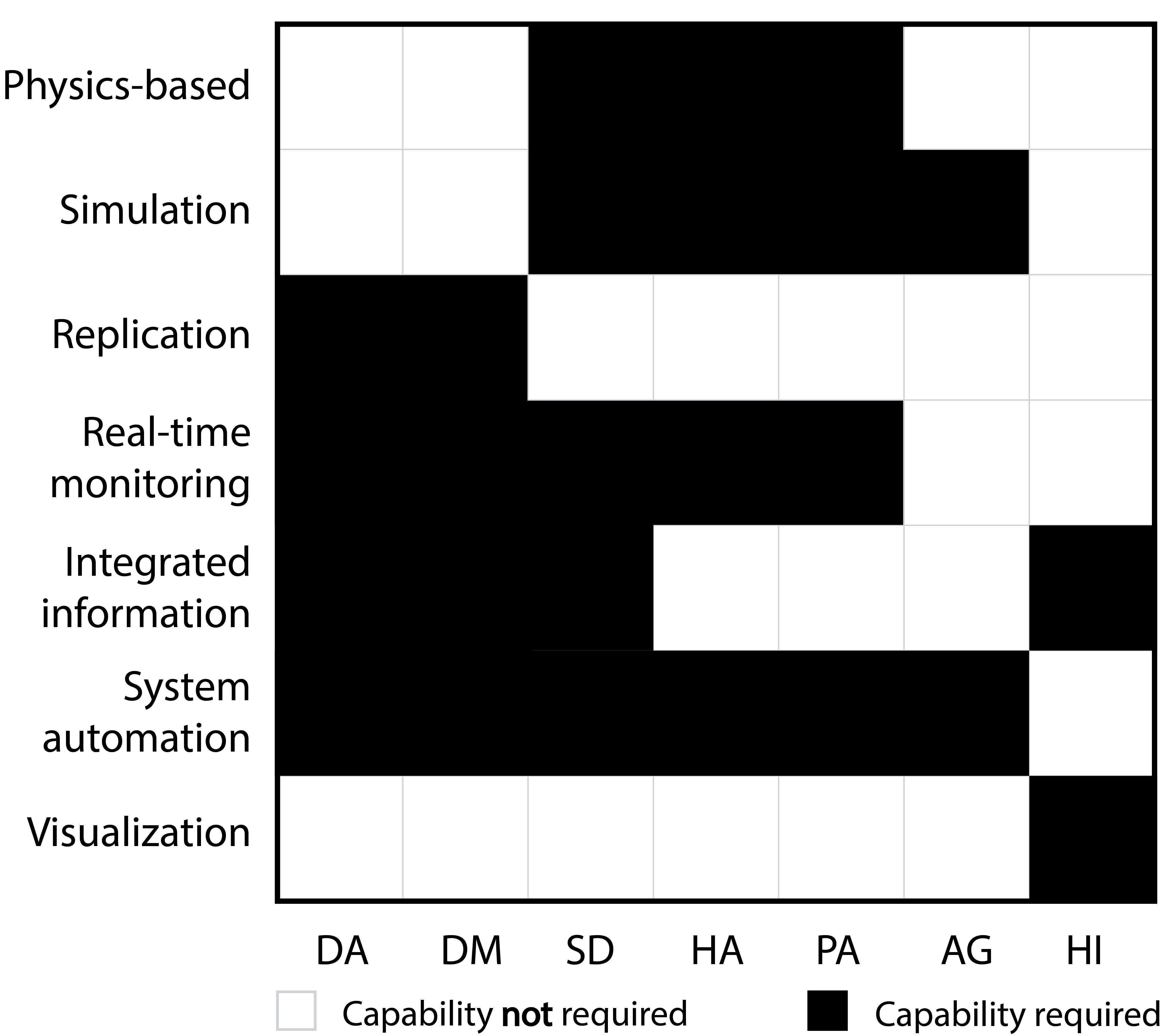}
\caption{Capability grid illustrating the relationships between PMx modules and the DT capabilities they support.}
\label{fig:Figure5}
\end{figure}

In Section~\ref{sec:Section2}, we outlined the existing standardized modules that support PMx tasks, and in Section~\ref{sec:Section3}, we identified the DT capabilities that facilitate these tasks. These two dimensions form the axes of the capability grid depicted in Fig.~\ref{fig:Figure5}. At each intersection on the grid, the presence or absence of a specific DT capability (represented on the $y$-axis) required by a particular PMx module (represented on the $x$-axis) is indicated. The value at each intersection is binary: either the capability is required or it is not. A relationship between a specific DT capability and a PMx module is considered established if more than 50\% of the selected survey articles emphasize its importance. This highlights the specific demands of each module-DT pairing. While it may be clear which capabilities are required for different modules, the real challenge -- and the area ripe for standardization -- lies in the complexities within each identified area on the grid. This pertains to the requirements and methodologies for implementing certain capabilities to enable scalable and automated operations within individual modules or across the entire PMx system.

\subsubsection{Extracting Requirements from the Capability Grid} 
\label{sec:Section4-A-2}
IRs are defined as specifications of an informational need, while FRs are defined as specifications of behaviors that the system must offer to enable users to achieve their goals~\cite{sommerville2009deriving,shah1988functional}. Although existing literature and surveys on PMx provide comprehensive perspectives on systematic design methodologies for various types of equipment~\cite{ran2019survey,zhang2019data}, well-defined requirements (i.e., IRs and FRs) are not adequately addressed. Some studies discuss requirements at an application-specific level, such as information systems for heating, ventilation, and air conditioning (HVAC) systems~\cite{balaji2016information}, and functional and environmental requirements for aircraft-related PMx~\cite{li2020toward}. Others partially identify the requirements or suggest that they should be explored in future work~\cite{jimenez2019system,justus2018capability}, often leaving these discussions as afterthoughts.

As a result, the literature lacks a comprehensive and systematic perspective that thoroughly articulates IRs and FRs in the context of PMx systems. Instead of adopting an application-centric approach to identifying IRs and FRs, which limits the PMx review to a single application or area (e.g., aircraft, industrial equipment), this study employs a module-based (or `pillar-based') strategy using the three pillars of PMx introduced in Section~\ref{sec:Section2}. An application-specific approach may not provide exhaustive insights due to the potential scarcity of detailed, domain-specific studies in certain fields. The module-based approach classifies PMx tasks into three comprehensive categories: SD, HA, and PA, each representing a stage in the PMx process -- anomaly detection, diagnosis, and prognosis, respectively. This approach offers a more precise roadmap for investigating a wide range of applications, thereby facilitating a more comprehensive exploration of the relevant topics.

Drawing inspiration from the system and software industries~\cite{lana2021data,sommerville2009deriving}, an extensive literature survey is conducted within each pillar (and across different application areas) to identify the IRs and FRs that support PMx systems. For example, physical properties such as equipment dimensions and material characteristics (whether for an aircraft or a CNC machine) are commonly used in studies focusing on fault detection, diagnosis, and prognosis. As a result, the category of `physical properties' is identified as one of the key IRs essential for PMx system autonomy.

\subsubsection{Asset Diversity} 
\label{sec:Section4-A-3}
\begin{figure}
\centering
\includegraphics[width=1\linewidth]{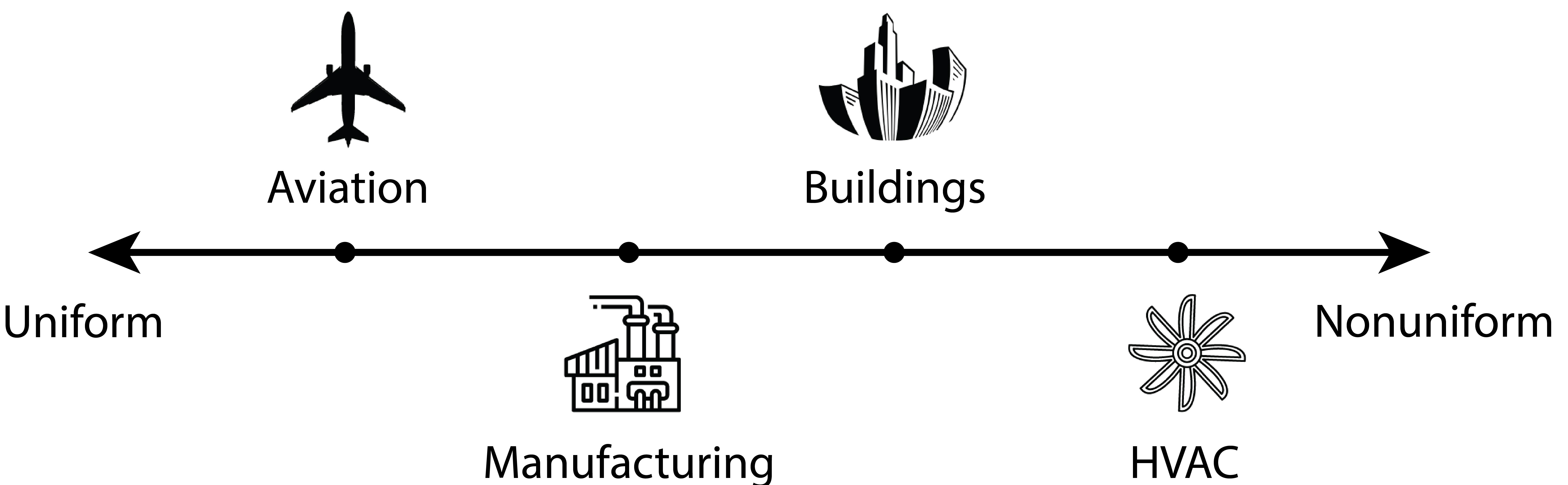}
\caption{Uniformity of assets across different PMx applications.}
\label{fig:Figure6}
\end{figure}

The potential diversity of assets within any given industry poses a significant challenge to the uniform development and application of DT-driven PMx across assets. When assets are uniform, implementing a DT framework for PMx across an entire fleet can be straightforward. However, when assets are heterogeneous, DT-driven PMx activities must be tailored to each specific asset. The concept of uniformity refers to the ease of applying a DT framework designed for one asset across the entire fleet. For example, uniformly produced products like aircraft require less effort to deploy a scalable, high-fidelity simulation compared to structures~\cite{boje2020towards}. Manufacturing plants fall between these two extremes; while reference models are available~\cite{bevilacqua2020digital}, they are more susceptible to changes than those for aircraft. On the other hand, HVAC systems exhibit a high degree of heterogeneity due to environmental disturbances and the complex thermal dynamics of the buildings they serve~\cite{homod2022innovative}. Fig.~\ref{fig:Figure6} illustrates this spectrum by including examples from various industries, providing insights into which sectors face greater challenges in adopting PMx strategies on a larger scale.

In the following subsection, we discuss the IRs and FRs within the context of the spectrum's two extremes -- aviation (uniform) and HVAC (nonuniform) -- to help readers better understand and align their own applications along this spectrum. While aviation and HVAC are used as representative examples at the extremes of the uniformity spectrum, this approach is intended to offer a conceptual framework that all industries can apply to assess their specific circumstances and challenges. By covering these boundary conditions, we aim to enhance situational awareness for stakeholders across diverse industries, encouraging broad applicability and insightful comparisons rather than limiting the discussion to only those sectors detailed.

%%%%%%%%%%%%%%%%%%%%%%%
%%%%% SECTION 4 B %%%%%
%%%%%%%%%%%%%%%%%%%%%%%

\subsection{Informational Requirements} 
\label{sec:Section4-B}
 A total of six IRs are identified, covering the physical entities to be maintained, the reference values that gauge their states, the operational context in which these entities function, the performance metrics used to quantify their health and performance, the historical data used for diagnostics and prognostics, and the potential faults within these physical entities. Each IR plays a unique, yet interconnected, role in the overall functioning and effectiveness of PMx systems. The following sections explore each of these IRs, detailing and contextualizing their specific features, importance, and implementation within the aviation and HVAC domains, as discussed in Section~\ref{sec:Section4-A-3}. 

To ensure our standardization efforts have broad applicability across various PMx solutions, all IRs are derived from a general PMx context rather than exclusively from a DT-enabled PMx perspective. However, for readers specifically interested in how each requirement is represented or realized within DT-enabled PMx systems, we provide relevant references at the end of each requirement's discussion. For a more detailed analysis of the relationship between IRs for PMx tasks and DTs (e.g., models, environments), we conducted a comprehensive synthesis in a previous work~\cite{ma2024stateoftheart}.

\subsubsection{IR1: Physical Properties}
\label{sec:Section4-B-1}
Physical properties refer to the spatial information and characteristics of an asset or its components. These properties are crucial across various industries, capturing details about the structure, size, and layout of physical entities. The form of physical information can vary, with its exact representation typically determined by the specific industry in which it is used (e.g., aircraft, manufacturing). For a more in-depth understanding of how IR1 is leveraged within PMx DTs across diverse industries, readers are encouraged to consult the comprehensive explorations provided in the literature review~\cite{kapteyn2022data,kapteyn2020toward,ye2020digital}.

\textbf{Aviation:} The specific physical properties considered in aviation maintenance are largely dependent on the metrics used (e.g., workability, safety level) and the particular substructures in focus (e.g., powerplant, wings). For example, in fuselage maintenance, scholars utilize spatial attributes such as slenderness ratio and fuselage radius, along with material properties like density, tensile strength, and Young's modulus. These parameters are essential for determining safety levels and predicting the future state or RUL of the structure~\cite{weiland2020structural,yiwei2019model,wang2018predictive}. Similarly, in wing maintenance, researchers rely on information regarding wing configuration (e.g., length, width, material properties) and specific coordinates (e.g., apex location, airfoil) to assess the current or future condition of the structure~\cite{li2017dynamic,zeng2021prognosis}.

\textbf{HVAC:} Due to their nonuniformity, the physical properties of HVAC systems depend heavily on the specific substructures involved and their associated fault types. For instance, in chiller maintenance, spatial and property information is used at both the system and component levels, including the geometry of the suction and discharge valves, building envelope data (in terms of $x$-$y$-$z$ coordinates), and material properties~\cite{dong2014bim}. Alternatively, in damper maintenance, spatial and property information is primarily focused at the system level, such as the types of Air Handling Units (AHUs) and their locations and capacities~\cite{YU2014550}. Beyond these component-specific details, capturing the relationships between components using semantic representation is crucial in real-world applications~\cite{balaji2016information}.

\subsubsection{IR2: Reference Values}
\label{sec:Section4-B-2}
Reference values serve as benchmarks for determining when an asset or its components are likely to fail. These values are crucial for assessing the current and future states of a component and may include lifetime data, run-to-failure data, or threshold values. Due to the scarcity of data, threshold values are most commonly applied. In PMx tasks, reference values provide a comparative metric against current performance, enabling the early identification of deviations or anomalies that could indicate a potential failure. They are also essential for determining the RUL and potential failure modes of the component, thereby ensuring that timely maintenance actions can be taken. For a more detailed exploration of how IR2 is utilized within PMx DTs across various industries, readers are encouraged to consult the comprehensive discussions in the literature review~\cite{luo2020hybrid,moghadam2021digital,tzanis2020hybrid}.

\textbf{Aviation:} Reference values play a crucial role in the maintenance of aviation components such as the fuselage and wings. For instance, during fuselage maintenance, data related to critical elements of the fuselage (e.g., skin material, longitudinal lap joints) is used to estimate the reference signal or reference state~\cite{weiland2020structural,wang2018predictive}. In contrast, wing maintenance requires a more comprehensive set of reference values beyond just material properties. Variables such as maximum tensile strength and the dimensions of unacceptable flaws are commonly used in this context~\cite{yousuf2017prognostic}. This variation highlights the nuanced application of reference values across different sub-domains of aviation maintenance.

\textbf{HVAC:} Reference values in the context of HVAC system PMx are employed in various ways~\cite{katipamula2005i}, with a recent trend in the field being the adoption of learnable reference values. This shift is driven by the inherent complexity and nonuniformity of HVAC systems, which necessitates adaptive and dynamic reference models. For example, in the case of AHUs, fault-free states are often used as reference values. These values are employed to create calibrated models that serve as baselines, triggering anomaly detection or diagnostic processes when significant deviations are observed~\cite{ANDRIAMAMONJY2018508,MONTAZERI2020101388}. The learning mechanisms used to determine these reference values can vary. Some approaches leverage unsupervised learning techniques, establishing thresholds through Q-statistics or squared prediction errors~\cite{GUO2017744,DU20073221,balak2014hvac}, enabling the discrimination between fault-free and faulty component statuses. Alternatively, regression-based models may be used, where manually set thresholds or bounds detect deviations between current and faulty statuses~\cite{GAO201669,YANG2013132}. Regardless of the specific method, the overarching principle remains the same: deviations from learned reference values indicate potential system faults, facilitating early detection and remediation in HVAC systems.

\subsubsection{IR3: Contextual Information}
\label{sec:Section4-B-3}
Contextual information refers to data used to determine the current or future states of a component or to make decisions and recommendations without being directly derived from the physical entity being maintained. Examples include environmental data and temporal and spatial information from sensor measurements. Contextual information is crucial for understanding the external factors that may impact the performance, degradation, and lifespan of the physical entity. Additionally, low-level contextual data (e.g., temperature readings) can serve as input for deriving high-level contextual data (e.g., environmental thermal profiles). By incorporating both low-level and high-level contextual information into their analysis, organizations can gain a more comprehensive understanding of the factors affecting their assets, enabling more accurate predictions and informed maintenance decision making. For a more detailed exploration of how IR3 is utilized within PMx DTs across various industries, readers are encouraged to consult the comprehensive discussions in the literature review~\cite{TAO2018169,kapteyn2020toward,kapteyn2022data}.

\textbf{Aviation:} Contextual information supports various substructures within the aviation maintenance framework. For example, research on empennage maintenance often relies on environmental data, which is instrumental in determining the future state of the associated structure~\cite{montel2016validation}. In contrast, the maintenance of powerplant components requires a more extensive collection of contextual characteristics. These include a wide range of parameters such as the total temperature and pressure at the fan inlet, output from the low
pressure compressor, outlet of the high pressure compressor, and output from the low pressure turbine~\cite{VERMA20061342}. This comprehensive set of contextual information is critical for accurately assessing the current state of powerplant components and predicting their future performance and potential failures.

\textbf{HVAC:} For HVAC systems, contextual information is relatively consistent across different subcomponents. Two types of contextual information are commonly used in most HVAC-related PMx studies. The first type relates to the specific context of each measurement, including details such as the subject of the measurement (e.g., temperature) and the zone where the measurement is taken. The second type concerns the environmental context surrounding the system itself, encompassing external factors such as ambient temperature and humidity~\cite{haruehansapong4167802deep}, zone occupancy rates~\cite{LIN2020106505}, and carbon dioxide ($\mathrm{CO}_2$) concentration levels~\cite{behraven2018}. This consistent use of contextual information highlights its crucial role in HVAC systems' PMx tasks, providing essential inputs for PMx models and algorithms in this field.

\subsubsection{IR4: Performance Metrics}
\label{sec:Section4-B-4}
Performance metrics refer to the operational performance of an asset or its components. Various metrics have been developed and employed to quantify the performance of components in PMx systems. These metrics, which can be customized for specific applications or based on standardized industry measures, provide valuable insights for making informed maintenance decisions and scheduling appropriate maintenance activities. Customized metrics are tailored to the unique characteristics and requirements of a particular component, considering its specific operating conditions and potential failure modes. Examples of customized metrics might include wear rates for specific material combinations~\cite{woldman2015abrasive}, performance degradation curves for a unique machine design~\cite{aivaliotis2021industrial}, or efficiency loss rates under particular environmental conditions. Standardized metrics, on the other hand, are based on industry-wide benchmarks or accepted best practices. These metrics offer a consistent and comparable way to assess the performance of structures or components across different applications and industries. Examples of standardized metrics include the mean time between failures~\cite{duenckel2017preventive,hruz2020innovative}, mean time to repair~\cite{masood2019aiops}, and overall equipment effectiveness. For a more detailed exploration of how IR4 is utilized within PMx DTs across various industries, readers are encouraged to consult the comprehensive discussions in the literature review~\cite{xiong2021digital,roy2020digital,yu2021digital}.

\textbf{Aviation:} In powerplant maintenance, performance metrics are essential for representing the system's workability and efficiency. Customized metrics are tailored to specific operational parameters and characteristics and are used to predict the current or future state of the powerplant. This customization often involves a weighted sum of flight conditions (e.g., altitude, Mach number) and sensor measurements (e.g., fan speed, fan flow)~\cite{frederick2007user}. These metrics can be tied to either static thresholds\cite{xiong2021digital} or dynamic ones that adapt to changing conditions~\cite{KO2022116094}. Standardized metrics are also employed for similar predictive tasks. For instance, widely recognized measures such as propulsive efficiency or Froude Efficiency are used to assess system performance~\cite{national2016commercial}. The use of both customized and standardized metrics highlights the multifaceted approach required in powerplant maintenance to achieve comprehensive and accurate PMx tasks.

\textbf{HVAC:} The performance metrics for HVAC systems are predominantly classified as standardized metrics, used to quantify characteristics such as capacity and efficiency across various substructures (e.g., fan, coil). For example, cooling capacity can be measured by controlling the flow rate of chilled water~\cite{liang2007model}, while power efficiency can be determined by monitoring the power transferred to the flow relative to the power consumed by a fan or pump~\cite{rafati2022fault}. Beyond assessing the performance of individual substructures, these metrics can also represent the overall performance of the corresponding PMx system. This can be achieved through the use of bespoke performance indices~\cite{nathalie2017hvac} or metrics that adhere to industry standards~\cite{elnour2020sensor}. The application of performance metrics in HVAC systems underscores their fundamental role in enabling effective PMx activities.

\subsubsection{IR5: Historical Data}
\label{sec:Section4-B-5}
Historical data refers to the maintenance records and usage profiles of an asset or its components. This data plays a critical role in the PMx process, as it is often used to estimate the current or future state of the physical entity. These records may include information on previous repairs, component replacements, inspection results, and observed failure modes, among other relevant details. Analyzing this data can help identify trends and patterns in equipment performance and degradation, which in turn informs maintenance decisions and strategies. It can also aid in projecting future usage by examining past usage patterns and trends, along with insights from maintenance professionals and engineers, to predict future operating conditions and equipment performance. For a more detailed exploration of how IR5 is utilized within PMx DTs across various industries, readers are encouraged to consult the comprehensive discussions in the literature review~\cite{cohen2021smart,liu2018design,liu2018industrial}.

\textbf{Aviation:} In aviation applications, historical data is particularly crucial for diagnostics and prognostics. Maintenance records related to the aircraft's structure are often used to assess its current condition~\cite{verhagen2018predictive}. Operational data -- such as intervals between flights, flight duration, loading logs, and flight length -- is used to predict the aircraft's future usage profile~\cite{tuegel2022evaluation,de2022alarm}. These predictive profiles help formulate maintenance decisions and scheduling~\cite{wang2018predictive,tuegel2022evaluation}. Additionally, with the growing trend of ML, historical data is increasingly used to project historical measurements onto quantities of interest, such as fault type or health index~\cite{Daily2017}. As a result, historical data plays a pivotal role in enabling accurate diagnostics, prognostics, and maintenance decision-making processes within the aviation industry.

\textbf{HVAC:} Historical data serves two primary functions in the context of HVAC systems~\cite{katipamula2005i}. First, in data-driven methodologies, historical data from repositories forms the foundation for PMx systems. This information allows the systems to uncover interdependencies between components and develop projections from inputs to outputs based on correlations within the features~\cite{kim2018review}. By leveraging past patterns and trends, this data-driven approach enhances the predictive capabilities of the system. Second, historical data plays a significant role in forecasting the future context and usage profile of HVAC systems~\cite{WU2021110487}. By analyzing past usage trends and operational contexts, the system can make informed predictions about future usage patterns. This predictive power enables more efficient maintenance scheduling, optimized energy usage, and improved overall system performance. Thus, the use of historical data in HVAC system PMx is integral to both understanding past performance and informing future operations.

\subsubsection{IR6: Faults}
\label{sec:Section4-B-6}
A fault refers to any anomaly, malfunction, or failure of an asset or its components. For components requiring maintenance, multiple types of faults can occur, such as cracks and holes in machines or sensor errors. These faults may either be identified as anomalies that require further manual inspection or explicitly identified in terms of their location, severity, and associated maintenance recommendations. Faults can impact the performance, safety, and lifespan of equipment or components, making it essential to detect, diagnose, and address them promptly. Regardless of the nature of the faults, it is crucial to clearly define and classify them before any practical use of the PMx system. Clearly identifying fault types enhances understanding of the applicability and limitations of the methods and results presented. This clarity also facilitates more effective knowledge transfer and collaboration among industry stakeholders, enabling the development of more targeted and efficient maintenance strategies. For a more detailed exploration of how IR6 is utilized within PMx DTs across various industries, readers are encouraged to consult the comprehensive discussions in the literature review~\cite{short2019pumping,xiong2021digital,deebak2022twin}.

\textbf{Aviation:} The occurrence and nature of faults in aviation are specific to each subcomponent. For instance, the fuselage, wings, and empennage often encounter faults such as cracks and disjoints. These faults are typically represented implicitly, where the faults are identified from real-time signals, thereby necessitating appropriate signal processing techniques for their detection~\cite{weiland2020structural,zeng2021prognosis,tuegel2022evaluation}. While it has been proposed to adopt more explicit representation forms, such as photographic images for fault detection in aviation, the diversity of available datasets is currently insufficient to fully realize this vision~\cite{stanton2023pmx}. In contrast to the aforementioned substructures, powerplants, due to their intricate construction and operational complexity, give rise to a considerably more
diverse range of potential faults. Among these, pressure-related faults at different locations (e.g., oil pressure inconsistencies, anomalies in turbine pressure) are commonly observed~\cite{frederick2007user}. Despite the proposed vision, the primary focus of fault detection in this area remains on real-time signal processing techniques.

\textbf{HVAC:} The complexity of HVAC systems, especially in commercial buildings, results in a diverse array of fault types,
ranging from issues related to AHUs and Variable Air Volume (VAV) boxes to those related to sensors and actuators. In contrast
to aircraft, the extensive availability of HVAC systems has allowed for a more thorough study of potential faults, resulting
in the creation of exhaustive lists detailing the diverse range of faults that can occur in various HVAC
subcomponents~\cite{CHEN2022112395}. Due to the exhaustive nature of faults in the HVAC field, the PMx pipeline in HVAC can be
identified as a three-step process related to faults, which comprises fault detection, fault isolation, and fault
identification~\cite{katipamula2005i}. The survey also suggested that recent PMx tasks for HVAC systems are predominantly
related to AHU and chiller problems~\cite{CHEN2022112395}. These issues, along with sensor malfunctions in HVAC systems, vapor
compression systems, and variable refrigerant flow systems, collectively account for more than 80\% of the HVAC Fault
Detection and Diagnosis (FDD) challenges. Faults related to AHU could be attributed to various components of it, each of which
has specific conditions and performance metrics that can indicate potential failures. In the case of chillers, two types of
faults have high frequencies and are vastly studied in the literature. The first type pertains to issues with the primary
fluid (i.e.,~refrigerant), such as refrigerant leaks~\cite{HAN2019540}. The second type is related to the secondary fluid
(i.e.,~water), such as reduced evaporator water flow~\cite{yan2018cost,he2016fault}. In most cases, these faults are detected
through the implicit information embedded within the system's signals, necessitating advanced data processing and analytical
techniques. Unlike aviation where high-resolution images are easily available due to the external visibility of some critical
components, HVAC systems primarily comprise internal components housed within enclosures, thus limiting the availability of
high-quality images for explicit fault detection. However, there is a growing interest in adopting more explicit
representations, similar to the trends observed in aviation, to enhance fault detection in HVAC systems. Such a trend could
involve the use of infrared imaging for identifying abnormal equipment and pipe temperatures or employing video imaging to
monitor frost levels~\cite{yang2023cv}. Still, the development and availability of the necessary datasets for such
advancements are yet to be fully realized.

%%%%%%%%%%%%%%%%%%%%%%%
%%%%% SECTION 4 C %%%%%
%%%%%%%%%%%%%%%%%%%%%%%

\subsection{Functional Requirements}
\label{sec:Section4-C}
After comprehensively identifying all IRs, we now shift our focus to delineating the FRs. The design and operation of PMx systems involve a set of foundational FRs that are essential for effective and reliable operation across various contexts. These systems must maintain consistency with physical theories and be aware of the operational context. Interpretability is crucial to making system decisions understandable to human operators. The systems should be robust and adaptive to handle varying conditions and environmental changes. Scalability and transferability are also vital to accommodate different workloads and domains. Lastly, acknowledging uncertainty is necessary for the system to recognize and manage uncertain conditions and data. 

The following sections explore each of these FRs, detailing their specific features, importance, and implementation within the aviation and HVAC domains, as discussed in Section~\ref{sec:Section4-A-3}. To ensure our standardization efforts have broad applicability across various PMx solutions, all FRs are derived from a general PMx context rather than exclusively a DT-enabled PMx context. However, for readers specifically interested in how each requirement is represented or realized within DT-enabled PMx systems, we provide relevant references at the end of each requirement's discussion. For a more detailed analysis of the relationship between FRs for PMx tasks and DTs (e.g., models, environments), we conducted a comprehensive synthesis in one of our previous works~\cite{ma2024stateoftheart}.

\subsubsection{FR1: Theory Awareness}
\label{sec:Section4-C-1}
Theory awareness refers to a system's ability to maintain efficient and consistent representations of the physical phenomena involved in the assets' operation~\cite{lana2021data}. Around the 2000s, PMx models were often expressed using standard mathematical functions, but they were limited by the lack of complexity and adaptability to new data or changes in the underlying system due to limited computing power and data resources~\cite{venkatasubramanian2003review,byington2002prognostic}. As other researchers have noted~\cite{lana2021data}, when AI was first introduced in various engineering fields, model development and evaluation diversified, often adapting to different technologies and algorithms without properly integrating domain knowledge. This made many models difficult for engineers and maintenance staff to apply effectively. Only recently has the implicit integration of domain knowledge, such as incorporating it into the boundary conditions or loss functions of data-driven models in PMx, shown the potential for accurate and realistic modeling of operational scenarios~\cite{yucesan2020physics,das2020data}. These functional requirements have demonstrated significant potential for enabling insightful learning across different application areas. For a more detailed exploration of how FR1 is utilized within PMx DTs across various industries, readers are encouraged to consult the comprehensive discussions in the literature review~\cite{ye2020digital,tzanis2020hybrid,luo2020hybrid}.

\textbf{Aviation:} In the early stages of integrating theory into AI-guided aviation maintenance, researchers incorporated theoretical components into their data-driven models. This was often achieved through model components (e.g., loss functions, physics/theory layers) or as boundary conditions~\cite{nascimento2019fleet,de2020physics}. Given that the theoretical aspects of aviation maintenance are well-studied, more recent work has focused on explicitly representing the physics associated with specific subcomponents. A common approach involves using hybrid models that combine a data-driven component, which primarily learns patterns and projections from data, with a physics/knowledge/theory-driven component that ensures the practicality of the model's output. Examples of such hybrid models include the aircraft structural model developed by the Air Force Research Laboratory. In these models, physics-based models work in parallel with data-driven models or operate in a cascade. For instance, a physics-based model provides the necessary framework and understanding of an engine's dynamics, which is then complemented by data-driven dynamic neural networks for effective fault diagnosis~\cite{SADOUGHVANINI2014234}.

\textbf{HVAC:} In the HVAC field, theories have been employed for several decades, generally divided into quantitative models, which use mathematical relationships based on process physics, and qualitative models, which are based on interpretative relationships derived from the same physics knowledge~\cite{katipamula2005i,katipamula2005ii}. Recently, researchers have developed numerous methods to combine knowledge-based and data-driven approaches, leveraging the strengths of both for HVAC-related PMx. Unlike in aviation, a significant body of research in HVAC explicitly integrates domain knowledge with data-driven methodologies, such as the AHU coil process model~\cite{yang2011novel}, the energy flow model~\cite{wu2011cross}, and Bayesian networks that map relationships between chillers and sensor features~\cite{wang2018enhanced}. These knowledge-guided systems typically pair with one of two types of data-driven models: statistics-based models and learning models. The application of domain knowledge in these hybrid models not only helps maintain consistent physical representations but also leads to significantly higher sampling efficiency, often requiring fewer key sensor data points (typically fewer than 10)~\cite{CHEN2022112395}.

\subsubsection{FR2: Context Awareness}
\label{sec:Section4-C-2}
Context awareness refers to the ability of the system to perceive and adapt to various operational and environmental
factors~\cite{GALAR2015137}. Even similar assets or components can exhibit variations in operational modes, data availability,
resource access, and environmental conditions due to the differing contexts in which they are deployed. In other
fields working towards system autonomy, such as autonomous vehicles, the pivotal aspect facilitating the shift from automation
to autonomy is the system's ability to be aware of its application context~\cite{TaxonomyAD}. Autonomy is particularly
important in the PMx domain because the system will be more efficient and ready for scaling, making the integration and
awareness of contextual information fundamentally critical for the effective operation and versatility of a PMx system. For a
more in-depth understanding of how FR2 is leveraged within PMx DTs across diverse industries,
readers are suggested to consult the comprehensive explorations detailed in the literature
review~\cite{roy2020digital,TAO2018169,werner2019twin}.

\textbf{Aviation:} In aviation PMx, context awareness primarily pertains to the operational and environmental parameters under
which the aircraft functions. Various maintenance methods are employed to handle different status and external conditions,
including but not limited to, altitude, ambient temperature and pressure~\cite{wang2020aircraft}, dust level and
humidity~\cite{bousdekis2019predictive}, wind speed~\cite{xiong2021digital}, and weather phenomena like rain or
ice~\cite{riba2021line}. These methods typically incorporate either learning models or multiple physics-based models suited to
diverse conditions. Notably, the aviation field faces comparatively harsher operating conditions, making context awareness a
vital element for establishing environmental requisites. Such requirements address variable environmental factors by defining
explicit constraints over them, for instance, maintaining temperature variation within a specific limit in the hosting
facility for Core System nodes~\cite{li2020toward}. This focus on context awareness underscores the critical role of adaptive
maintenance strategies in ensuring the reliability and performance of aviation systems in fluctuating environmental conditions
and various operating statuses.

\textbf{HVAC:} Unlike aviation, the interactions between context and assets are bidirectional. Faults in HVAC systems can
significantly impact indoor thermal conditions, such as humidity levels~\cite{CHEN2022112395}. Given that HVAC control systems
often require contextual data, they can be adversely affected by harsh outdoor environments characterized by fluctuations in
temperature and humidity. Therefore, in contrast to aviation, the efficacy of FDD in HVAC systems is influenced by the
relationship between the model and the context. Given the bidirectional interactions in HVAC systems, context awareness in
this field considers a broad spectrum of factors. These include not only environmental variables, such as outdoor temperature
and humidity, but also operational status like occupancy rate and seasonal operating rules. While environmental factors often
play a direct role in the system's physical condition~\cite{rijal2021development,elnaklah2021moving}, operational status can
indirectly influence the system's functioning by dictating its workload and operation schedules. Nevertheless, the greatest
challenge lies in identifying key contextual information and translating it into a machine-readable format. Agreeing on how
this connection is established is a complex task. Meanwhile, current research often addresses specific problems by observing
primary contextual factors in buildings directly without drawing links to secondary factors~\cite{hosseini2021knowledge}. This
underscores the need for more nuanced context modeling in HVAC systems to improve FDD effectiveness and enhance overall system
performance.

\subsubsection{FR3: Interpretability}
\label{sec:Section4-C-3}
Interpretability refers to the ability of the system to generate human-interpretable outputs. In the field of PMx, the quality
of decisions and actions produced drives the safe and economical operation of certain
equipment~\cite{vollert2021interpretable}. Despite technology advances, we still need to understand why systems
generate certain decisions and actions~\cite{arrieta2020explainable}. Although explainability is a broader concept that
includes more general transparency features of models, our focus remains primarily on interpretability because it directly
enhances the usability and effectiveness of PMx technologies by making complex systems comprehensible and actionable for
users. The concept of interpretability in PMx is often divided into two distinct aspects: local and global interpretability.
Local interpretability pertains to understanding the model's behavior for specific instances or inputs. In contrast, global
interpretability refers to comprehending the model's behavior on an aggregate level or as a whole. In the context of PMx,
local interpretability is generally given more weight due to the nature of the tasks, which often involve diagnosing and
forecasting issues for individual systems or components. This focus on specific instances, particularly outliers and abnormal
instances, makes local interpretability more meaningful and relevant for PMx. Hence, it is more frequently discussed in the
literature. The ability to explain why a model predicts a specific failure or outputs a specific recommendation, decision, or
action for a specific component at a specific time is invaluable for effective diagnosis and repair planning, thus
demonstrating the significance of local interpretability in PMx systems. For a more in-depth understanding of how FR3 is leveraged within PMx DTs across diverse industries, readers are suggested to consult the comprehensive
explorations detailed in the literature review~\cite{li2017dynamic,yu2021digital,kapteyn2020toward}.

\textbf{Aviation:} In aviation-related PMx, explaining the outputs typically follows two distinct strategies. The first
strategy, known as model-agnostic explanation, separates the task of explanation from the underlying ML model. In other words,
the model and its explanations are decoupled. Examples of this approach include the use of SHapley Additive exPlanations
(SHAP) values, which provide insight into the contribution of each input feature to the output~\cite{yang2022data}, and Local
Interpretable Model-agnostic Explanations (LIME), which trains local surrogate models to explain individual
predictions~\cite{essay75381}. Despite most studies in aviation PMx using model-agnostic explanation focusing on local
interpretability, some studies have explored global model-agnostic interpretability, such as using visualization techniques to
understand what a model has learned~\cite{forest2020large}. The primary advantage of these decoupled approaches is their
flexibility; system developers can employ any ML model to generate outputs, as the interpretive methods are model-agnostic.
The second strategy involves the use of intrinsically interpretable models. Linear or rule-based models were employed to
fulfill interpretability~\cite{lee2018interpretable,ishibashi2013gfrbs}. However, these methods are often subject to their own
disadvantages, such as difficulty modeling linear relationships for decision trees or nonlinearity or interaction having to be
hand-crafted for linear regression~\cite{molnar2020interpretable}. The lack of adaptability is also another potential drawback
as the opposite side of model-agnostic methods. 

\textbf{HVAC:} The concept of interpretability has only recently begun to be recognized as a crucial factor in evaluating research in the field of HVAC systems PMx, as highlighted in several reviews~\cite{CHEN2022112395,meas2022explainability}. Given the inherent complexity of modern HVAC systems and the unique designs discussed in Section~\ref{sec:Section4-A-3}, the models used for PMx tasks often lack the
ability to provide intrinsic explanations. This inherent limitation has led researchers to focus on
model-agnostic methods for explaining the fault classification results of ML models. Examples of these
include the use of LIME~\cite{srinivasan2021explainable,madhikermi2019explainable}, SHAP~\cite{belikov2022explainable}, and
visualization method via gradient-based localization~\cite{li2021explainable}. Each of these methods provides a means of
extracting interpretable insights from the predictions of complex, black-box models, thereby enhancing the accessibility and
usability of PMx in HVAC systems. Notably, as in aviation applications, the focus in HVAC systems PMx is predominantly on local interpretability. This focus reflects the practical need to understand and explain individual predictions, especially those related to faults and anomalies, rather than the broader behavior of the model. The emphasis on local interpretability underscores the value of actionable insights in the PMx field, as these localized explanations can directly inform targeted maintenance and intervention strategies.

\subsubsection{FR4: Robustness}
\label{sec:Section4-C-4}
Robustness refers to a system's ability to maintain acceptable performance under potential disturbances in both
physical and digital domains~\cite{lana2021data}. In real-world PMx scenarios, various communication challenges due to changes
in physical environments can disrupt the normal operation of the system or model. These challenges include, but are not
limited to, data acquisition errors and input/output errors between modules, as discussed in
Section~\ref{sec:Section2}~\cite{EFTHYMIOU2012221}. However, these considerations are often overlooked in research-level designs,
leading to the devaluation of the system's resilience to external disruptions. Beyond the internal errors within the PMx
system, performance can be significantly affected when data falls outside of the model's operational boundary, necessitating
extrapolation. In such situations, the model is expected to make predictions or inferences beyond its training data, a task
that is fraught with uncertainty and risk. Unfortunately, many research-level designs do not adequately address this
challenge. Univariate fault detection extrapolation methods for predicting maintenance have been found to be suboptimal,
lacking robustness and long-term maintainability~\cite{butte2018machine}. There is a crucial need for PMx systems to be
designed with a higher level of robustness and resilience, capable of handling both internal errors and extrapolation
challenges. For a more in-depth understanding of how FR4 is leveraged within PMx DTs across diverse
industries, readers are suggested to consult the comprehensive explorations detailed in the literature
review~\cite{kapteyn2022data,kapteyn2020toward,short2019pumping,moghadam2021digital}.

\textbf{Aviation:} While the robustness of communication errors within PMx systems remains somewhat underexplored in aviation, a few studies have addressed related extrapolation challenges. These efforts have employed either unimodal or multimodal approaches. For example, a unimodal strategy might involve using a re-scaled long short-term memory algorithm to address the issue of imbalanced datasets, a common concern in PMx models for aircraft component replacement~\cite{dangut2020rescaled}. Alternatively, a multimodal strategy might involve expanding the operational boundary directly, thereby enhancing robustness~\cite{vianna2017predictive}. These initiatives suggest potential avenues for further advancements in this critical aspect of PMx system development.

\textbf{HVAC:} For HVAC PMx applications, robustness in the context of sensor errors and extrapolation challenges has been
extensively studied, with the former receiving attention for nearly two decades. The International Energy Agency has
specifically highlighted the significance of measurement errors, stating: \textit{``measurement errors are a major obstacle to the
successful application of FDD tools in HVAC systems}~\cite{dexter2001demonstrating}.'' While methods, such as Principal
Component Analysis (PCA)~\cite{wang2006robust}, artificial neural networks~\cite{du2014fault}, and probabilistic graphical
models~\cite{aniket2016robust} have been used to counter sensor bias or missing value errors, the development of FDD systems
that can accommodate a variety of sensor errors remains an underexplored area. Looking at extrapolation challenges, the
significant diversity of building energy systems, as outlined in  Section~\ref{sec:Section4-A-3}, implies that a higher degree of
accuracy for PMx tasks on the training dataset could potentially increase the risk of overfitting. This, in turn, could
compromise reliability and robustness in real-world applications~\cite{zhao2019artificial}. Consequently, more recent studies
have begun to focus on improving not just accuracy but also reliability, robustness, and generalizability from both data and
model points of view~\cite{CHEN2022112395}. While significant strides have been made in the area of robustness in HVAC PMx
applications, there remains considerable scope for further exploration as described.

\subsubsection{FR5: Adaptivity}
\label{sec:Section4-C-5}
Adaptivity refers to the ability of the system to modify its internal process or behaviors based on the deterioration or
evolution of an asset. Unlike robustness, which focuses on the system's ability to maintain performance under unexpected
disruptions or errors, adaptivity emphasizes the system's flexibility in adjusting its internal processes or behaviors in
response to changes in the asset's state. In real-world PMx scenarios, adaptivity is crucial due to the non-static nature of
operational environments where equipment conditions and performance parameters are continually
changing~\cite{zenisek2019machine}. These dynamic conditions necessitate models that can update their understanding and
predictions based on new data, acknowledging the evolving nature of assets. In the field of ML, such variations of
relationships between inputs and outputs are known as concept drift, and it is, at present, an active research
topic~\cite{gama2014survey}. To effectively contend with the challenges posed by concept drift, there are two cascading
sub-requirements that demand consideration~\cite{lana2021data}: \textbf{(FR5-1)} the detection of concept drift and
\textbf{(FR5-2)} the adaptation process towards concept drift. However, existing research on adaptable PMx primarily centers
around model adaptability in response to concept drift, addressing FR5-2, while the exploration of FR5-1 remains largely
underexplored. For a more in-depth understanding of how FR5 is leveraged within PMx DTs across diverse
industries, readers are suggested to consult the comprehensive explorations detailed in the literature
review~\cite{luo2020hybrid,tygesen2019state}.

\textbf{Aviation:} Aviation-related PMx is a subfield that demands significant adaptivity. Under typical circumstances, if the degradation process of a physical entity follows a monotonic path, the relationship between the degradation and the model can be straightforwardly established~\cite{CORBETTA2018305}. However, when this is not the case, a common approach is to assume a linear degradation trajectory and implement a Wiener process. Despite this, the degradation of many critical aviation substructures, such as wings and engines, is often neither monotonic nor linear~\cite{legresley2001investigation}. To address nonlinear stochastic degradation, researchers frequently incorporate time-related parameters into their models. Examples include time-dependent drift coefficients~\cite{si2015adaptive}, time-dependent forgetting factors~\cite{berghout2020aircraft}, or dynamic sensor consistency metrics~\cite{zhang2020aircraft}, which are designed to complement learning models and accommodate nonlinearity in degradation paths through timely piecewise adaptation.

\textbf{HVAC:} The intricate nature of HVAC systems, compounded by the dynamic attributes of their corresponding environments,
necessitates the capability to handle concept drift. Without this, FDD algorithms risk being error-prone and
unreliable~\cite{wang2021fault}. However, as summarized by Chen et. al. ~\cite{CHEN2022112395}, a mere 15\% of the current papers
demonstrated adaptability, and methods that require only fault-free data are intrinsically more adaptive as faulty data might
be unavailable after concept drift. Examples of these methods include hybrid models~\cite{wang2021fault} and unsupervised
learning~\cite{hu2012chiller,balak2014hvac}. While adaptability in HVAC PMx is crucial, current literature reveals it to
be underrepresented, with a limited range of methods demonstrating this characteristic.

\subsubsection{FR6: Scalability}
\label{sec:Section4-C-6}
Scalability refers to a system's ability to maintain performance across a diverse range of workloads or be extended to varying scales. PMx research typically operates on a restricted scale, limited by the data volume suitable for experimental purposes. This contrasts sharply with the diverse and evolving demands of industrial systems, which can range from small, localized setups to expansive, interconnected networks~\cite{zhang2019data}. In practice, a PMx system should serve as a scalable solution, capable of managing either a large fleet or individual instances based on stakeholders' requirements. Scalability presents two opposing challenges. First, as the number of assets in a PMx task increases, the volume of data logs and performance data generated can grow significantly. Preparing sufficient computational resources for this data management can be both costly and unsustainable~\cite{sakr2016towards}. To address this, AI can be deployed to manage and utilize the data more efficiently~\cite{bandari2021comprehensive}. Second, when PMx tasks are required for individual assets using edge devices, the system's ability to maintain accuracy with limited data and computing resources becomes critical. However, this aspect remains under-researched in the literature. Notably, the application of a PMx system is closely tied to its scalability. As in many other engineering fields~\cite{lana2021data}, some PMx tasks are more amenable to scaling due to the nature of the problem and system design, while certain PMx subfields exhibit high sensitivity to scalability~\cite{gigoni2019scalable}. Consequently, there is a strong motivation for research investigating scalability within PMx systems. For a more detailed exploration of how FR6 is leveraged within PMx DTs across various industries, readers are encouraged to consult the comprehensive discussions in the literature review~\cite{dhada2021deploy,rossini2020replica}.

\textbf{Aviation:} Advancements in IoT and edge computing have led to recent efforts to study both of the aforementioned
scalability challenges within the context of aviation PMx. The first challenge, related to the inability of traditional
real-time and local monitoring techniques to collect aircraft telemetry from multiple data
sources~\cite{watkins2019prognostics}, has limited scalability within fleet management. To help address aspects of this
challenge, recent work has proposed leveraging IoT cloud-based services, which employ a single application instance serving
multiple clients, thereby facilitating scalable fleet management in the cloud~\cite{bouzidi2020efficient}. Advanced algorithms
and modeling alterations related to data mining, such as dimensionality reduction techniques and conventional classification
models, are being used to approach the second challenge, which is concerned with individual-level PMx
tasks~\cite{painter2006simulation,zahra2021}. Given the wide application of data-driven models and their own inductive biases,
the demand for scalability often transitions to the need for reliable and efficient models that can operate with limited
data~\cite{Kristensen2017PredictiveAW}. While the progressive application of IoT has paved the way for addressing the first
challenge of scalability in aviation-related PMx, the second challenge, maintaining efficiency and reliability at an
individual level, remains understudied.

\textbf{HVAC:} Given the inherent nonuniformity of HVAC systems outlined in  Section~\ref{sec:Section4-A-3}, the necessity for
scalable solutions is less prevalent in this field compared to aviation, primarily due to the impracticality of using a single
application instance for multiple disparate systems. Yet, it is imperative to recognize that scalable FDD systems hinge on the
judicious selection of models, as relying merely on developer preference or existing tools might suffice for lab-scale
endeavors but can falter in real-world, full-scale implementations~\cite{katipamula2005i}. The second challenge, on the other
hand, is practically meaningful but complicated by the heterogeneity in system design, spatial location, and occupancy
patterns~\cite{georgescu2017whole}. The applications of conventional ML methods are, therefore, incapable of producing
scalable solutions. A prevalent strategy in this domain involves the use of rule-based
models~\cite{chen2021development,house2001expert}. The fine-grained differentiation of rules facilitates the detection of
issues occurring within shorter duty cycles and the creation of more generalized rule sets, thereby fostering a more scalable
approach to system analysis.

\subsubsection{FR7: Transferability}
\label{sec:Section4-C-7}
Transferability refers to a system's capability to uphold its performance when deployed on assets or conditions distinct
from those on which it was initially trained~\cite{lana2021data}. Unlike scalability, which concerns the system's capacity to
handle diverse workloads or expand to different scales, transferability is concerned with the system's ability to maintain its
performance when applied to conditions or assets distinct from its training environment. It reflects the
challenge of applying a model, say trained on a helicopter, to different contexts or equipment, such as a conventional
civilian aircraft, underpinning the complexity of diverse real-world deployments. In many fields employing ML models, not
limited to PMx, the standard practice is to evaluate models using test data. However, little attention is typically paid to
the performance of these models in different contexts or when applied to data from other equipment. The ease or difficulty of
achieving transferability may vary depending on the specific source and target domain in different PMx applications. For
instance, for a model trained on the dataset collected from a helicopter, it is easier to focus on a target domain involving
another make of a helicopter than to deal with a conventional civilian aircraft that has a different operating mechanism.
Regardless of the difficulty in domain adaptation, transferability is crucial to PMx systems, as technical integration
manually can make the implementation of PMx solutions costly~\cite{shao2019transfer}. Transfer learning has made significant
strides in PMx-related tasks, garnering interest from academia and industry alike and emerging as a key research
area~\cite{LI2020121}. For a more in-depth understanding of how FR7 is leveraged within PMx DTs across
diverse industries, readers are suggested to consult the comprehensive explorations detailed in the literature
review~\cite{xu2019twin,deebak2022twin,kapteyn2022data}.

\textbf{Aviation:}  The significant discrepancies between simulated and real-world data, also known as the ``reality
gap''~\cite{wada2022sim}, is often larger in aviation than in many other PMx applications, such as manufacturing and
healthcare. This fact emphasizes the necessity of transferability as an indispensable characteristic of aviation-related PMx
systems. Two distinct methodologies were found in the literature. The first approach generates simulations with the
interdependence of degradation and operating conditions~\cite{gardner2022population,arias2021aircraft}, often through the
establishment of a metric quantifying the distance between two feature spaces, either raw or processed. The second approach
explicitly models ``transferability'' as a quantifiable characteristic, augmenting it through deep transfer learning
techniques~\cite{dong2019implementing}. In essence, the overarching goal of these methodologies is to bridge the gap between
simulated and real-world scenarios, enhancing the applicability and generalizability of PMx in aviation.

\textbf{HVAC:} Notably, for nearly two decades, the emphasis has been on the need for tools that seamlessly integrate into the
working environments of building operators and maintenance service providers. This call to action arises from the realization
that many FDD methods have been primarily validated in controlled settings like laboratories or special test
environments~\cite{katipamula2005ii}. Although there have been strides in advancing transferability within the HVAC-related
PMx, the majority of these attempts have been confined to specific settings, such as simulations or experimental data
pertaining to a particular type of HVAC system under designated operating conditions~\cite{CHEN2022112395}. Given the inherent
nonuniformity of HVAC systems outlined in Section~\ref{sec:Section4-A-3}, it is difficult, if not impractical, to produce universal
applicability and transferability encompassing different HVAC settings explicitly. On the other hand, recent studies started
to utilize transferability implicitly. By investigating the transferability of faults across varying building operation
conditions, researchers seek to understand the interconnectedness of different operational components, the combined impact of
simultaneous faults, and the fluctuating severity of faults~\cite{ZHANG2017178}. Assuming the fault data is transferable, it
could pave the way to create a central dataset comprising collected fault data, thereby supporting online data-driven FDD
proposed in aviation applications~\cite{kapteyn2022data}. This represents a significant step towards optimizing the
generalizability and practical applicability of FDD methods in real-world HVAC systems.

\subsubsection{FR8: Uncertainity Awareness}
\label{sec:Section4-C-8}
Uncertainty awareness refers to a system's ability to recognize and quantify uncertainty inherent in its input, modeled processes, and outputs. In the context of data-driven models, uncertainty often arises due to the variability and
incompleteness of the data used for training these models. The field of PMx often encounters uncertainty, attributable to the
input data or environmental influences~\cite{van2013dynamic}. The convergence of these two sources of uncertainty leaves an
indelible mark on the outputs of these models. A failure to acknowledge such inherent randomness could precipitate unforeseen
consequences; therefore, it necessitates explicit assessment. For a more in-depth understanding of how FR8 is leveraged within PMx DTs across diverse industries, readers are suggested to consult the comprehensive
explorations detailed in the literature review~\cite{tygesen2019state,kaul2019digital,li2017dynamic}.

\textbf{Aviation:} The unpredictability inherent in faults and events pertaining to aviation safety incidents is often modeled
by experts drawing upon their accumulated field experience~\cite{kandel1991fuzzy,dym1991knowledge}. These models typically
take the form of a set of inference rules (for example, \textbf{if} premise \textbf{then} conclusion $\rightarrow$
\textbf{uncertainty}). However, the subjective nature of personal experience can lead to decisions and causal relationships
that are difficult to quantify. The advent of AI has introduced more robust ways of expressing this uncertainty. With the
explosion of data availability and the development of data mining techniques. One approach is to ``learn'' the rules and
membership functions directly from the data rather than relying solely on expert input, which is often static and not
adaptable. Existing literature employs fuzzy sets to represent the conceptual uncertainty (such as confidence level) of
various risks (e.g.,~improper maintenance or operator deficiency)~\cite{hadjimichael2009fuzzy}, as well as the degradation of
the subsystem of interest~\cite{lu2017multi}.

\textbf{HVAC:} In HVAC-related PMx systems, the main source of uncertainty comes from environmental factors. Environmental
factors are used in a vast range of PMx systems under different operational settings. Similar to the aviation field, fuzzy
inferencing is often utilized to guide this process~\cite{katipamula2005i,LO2007554}. However, considering that fuzzy
logic-related methodologies model uncertainties using logical operations that necessitate expert models -- particularly in
highly nonuniform HVAC systems -- the application of Bayesian probability to explicitly model and convey uncertainty has been
increasingly prevalent in recent studies. Existing research in this field has employed techniques such as Bayesian
inference~\cite{ng2020bayesian,li2021knowledge}, and unscented Kalman filtering~\cite{bonvini2014robust}.

%%%%%%%%%%%%%%%%%%%%%
%%%%% SECTION 5 %%%%%
%%%%%%%%%%%%%%%%%%%%%

\section{Identifying Key Research Gaps}
\label{sec:Section5}

With the IRs and FRs identified, the objective of this section is to conduct a thorough literature review spanning fields to determine the ways in which the IRs and FRs are currently being used within PMx DTs, enabling the identification, scoping, and prioritization of the key gaps and research areas needed to support a PMx DT framework.

%%%%%%%%%%%%%%%%%%%%%%%
%%%%% SECTION 5 A %%%%%
%%%%%%%%%%%%%%%%%%%%%%%

\begin{table}
\caption{Selection criteria for filtering abstracts.}
\setlength{\tabcolsep}{3pt}
\centering
\begin{tabular}{|p{40pt}|p{195pt}|}
\hline
Identifier&
Criterion\\
\hline
    A-CR1 & The study must be a primary study after 2012 in English\\
    A-CR2 & The study must clearly state the use of DTs to aid PMx tasks\\
    A-CR3 & The study must mention its aim (and, if any, results)\\
    A-CR4 & The study must not be a book, technical report, dissertation, or thesis\\
\hline
\end{tabular}
\label{Table3}
\end{table}

\subsection{Scoping the Current Predictive Maintenance Digital Twin Landscape}
\label{sec:Section5-A}
Following the same literature review procedure outlined in Section~\ref{sec:Section3}, the process of filtering relevant papers on DT-driven PMx comprises three steps: string search, abstract filtering, and content filtering. For the string search, we first define DT and PMx as root words and then broaden the scope to include words that have similar meanings across the literature. For example, since the term ``cyber-physical system'' is often used synonymously with DT, both terms are used in parallel. Similarly, PMx is used in parallel with CBM, prognostics and health management, diagnostics, and RUL. Additionally, since DT and PMx often function as application-oriented domains, we included terms related to the practical implementation of related technologies. Terms like ``case study'' and ``implementation'' are included to allow for the inclusion of works that not only discuss the theoretical aspects of PMx DTs, but also demonstrate their practical application in real-world scenarios. The final search query is as follows:

\begin{table}[h]
\caption{Selection criteria for filtering content.}
\setlength{\tabcolsep}{3pt}
\centering
\begin{tabular}{|p{40pt}|p{195pt}|}
\hline
Identifier&
Criterion\\
\hline
    C-CR1 & The study must define DT-related terms similar to the aforementioned selected surveys and reviews \cite{barricelli2019twin,fuller2020twin,JONES202036,KRITZINGER20181016,LIU2021346,qi2018digital}\\
    C-CR2 & The study must state its intention (e.g., framework, implementation, both)\\
    C-CR3 & The study must propose and address its own research question(s)\\
    C-CR4 & The study should state both its positive findings and negative findings \\
    C-CR5 & The study should clearly state challenges and open questions\\
    C-CR6 & The study must clearly define the assumptions, context, and design of its experiment\\
    C-CR7 & The study must have a traceable dataset (e.g., public dataset, self-collected dataset)\\
\hline
\end{tabular}
\label{Table4}
\end{table}

\begin{quote}
(``digital twin'' OR ``cyber-physical system'') AND (``predictive maintenance'' OR ``condition-based monitoring'' OR ``prognostics'' OR ``health management'' OR ``diagnostics'' OR ``remaining useful life'') AND (``case'' OR ``implementation'')
\end{quote}

\noindent Following this stage, a total of 583 manuscripts were earmarked for further analysis across Springer (148), ScienceDirect (94), ACM (119), IEEE Xplore (125), and others (97).

\begin{figure*}
\centering
\begin{subfigure}[b]{0.35\textwidth}
   \includegraphics[width=1\linewidth]{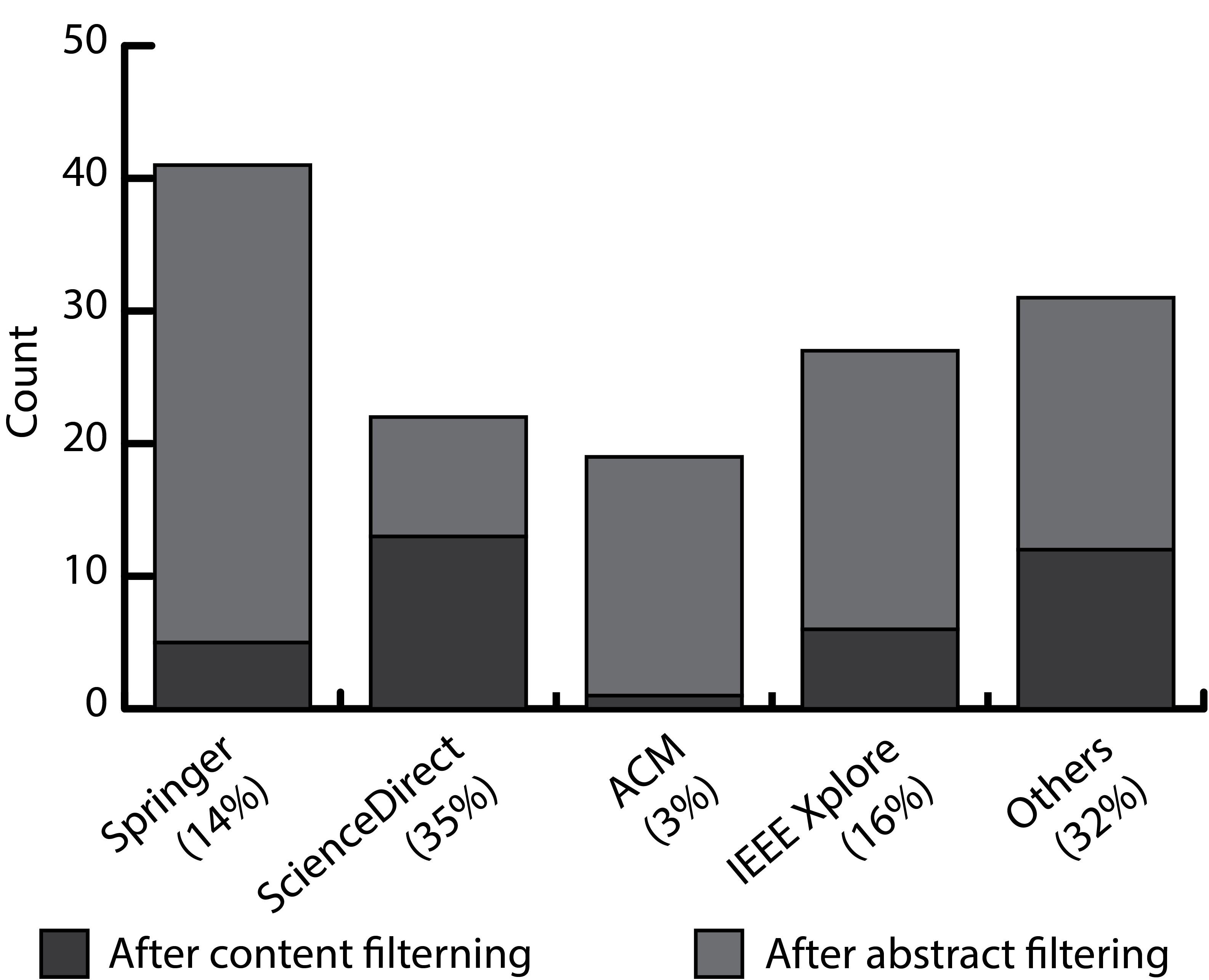}
   \caption{}
   \label{fig:Figure7a} 
\end{subfigure}
\hspace{1cm}
\begin{subfigure}[b]{0.35\textwidth}
   \includegraphics[width=1\linewidth]{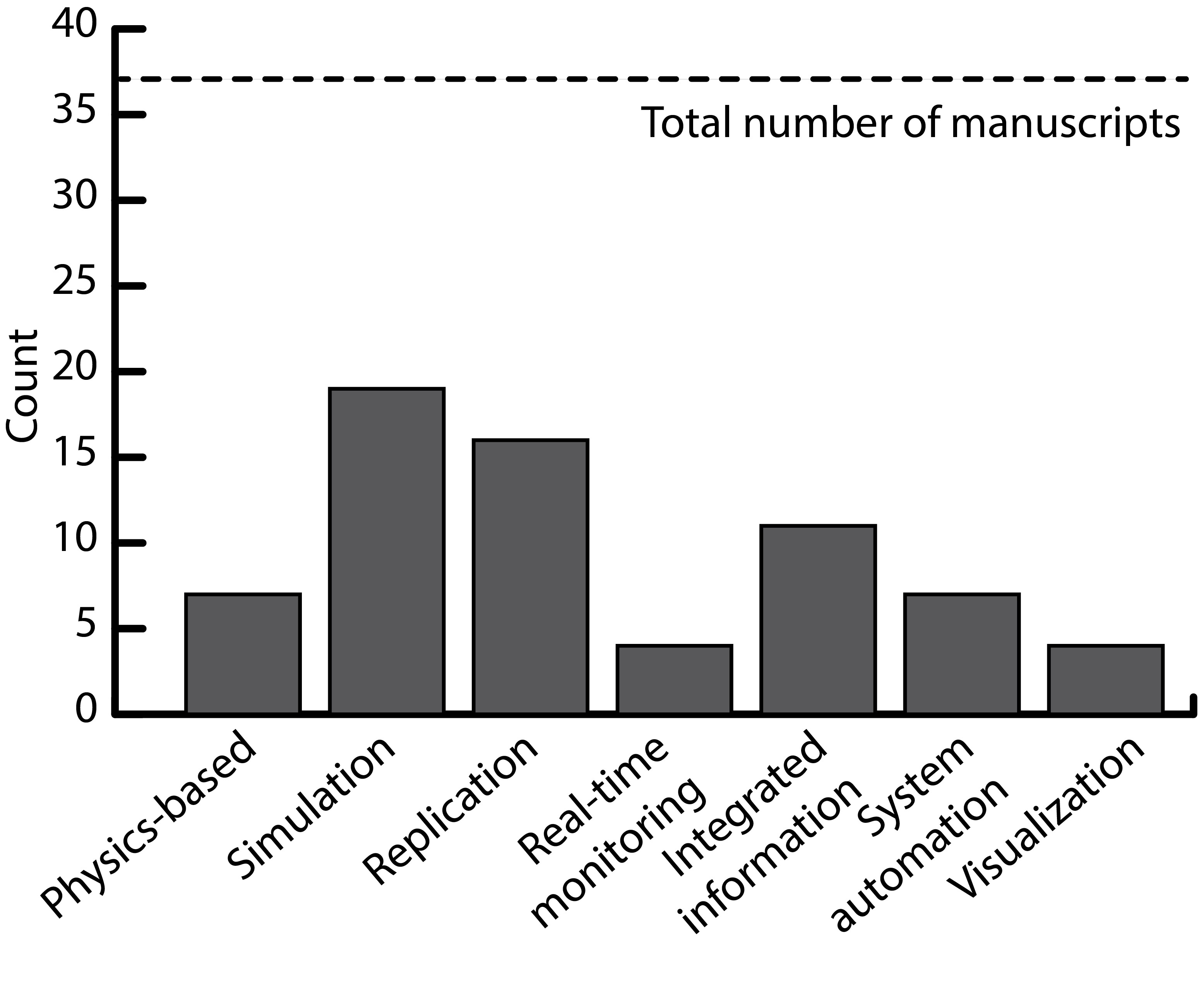}
   \caption{}
   \label{fig:Figure7b} 
\end{subfigure}
\hspace{1cm}
\begin{subfigure}[b]{0.35\textwidth}
   \includegraphics[width=1\linewidth]{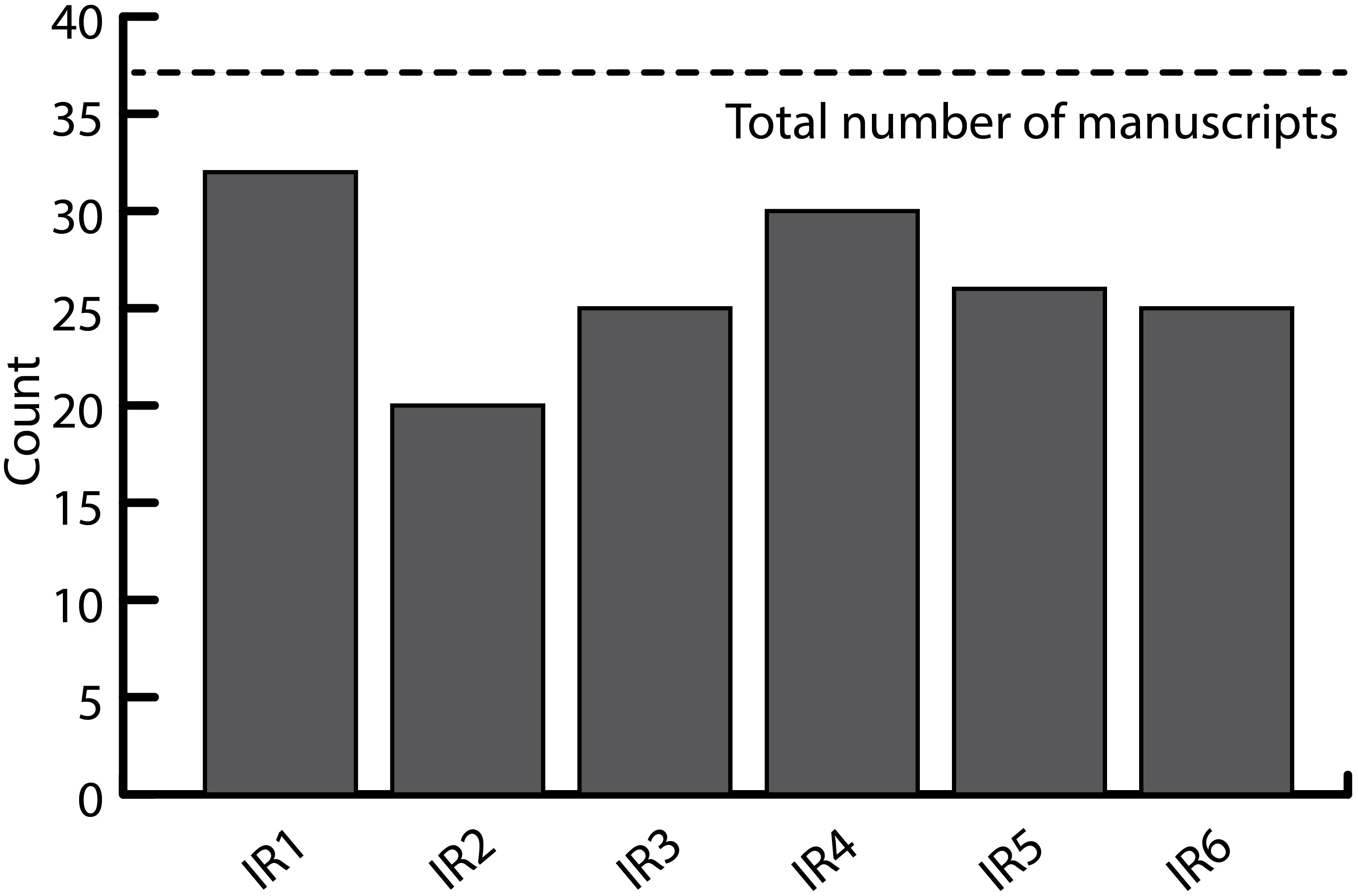}
   \caption{}
   \label{fig:Figure7c} 
\end{subfigure}
\hspace{1cm}
\begin{subfigure}[b]{0.35\textwidth}
   \includegraphics[width=1\linewidth]{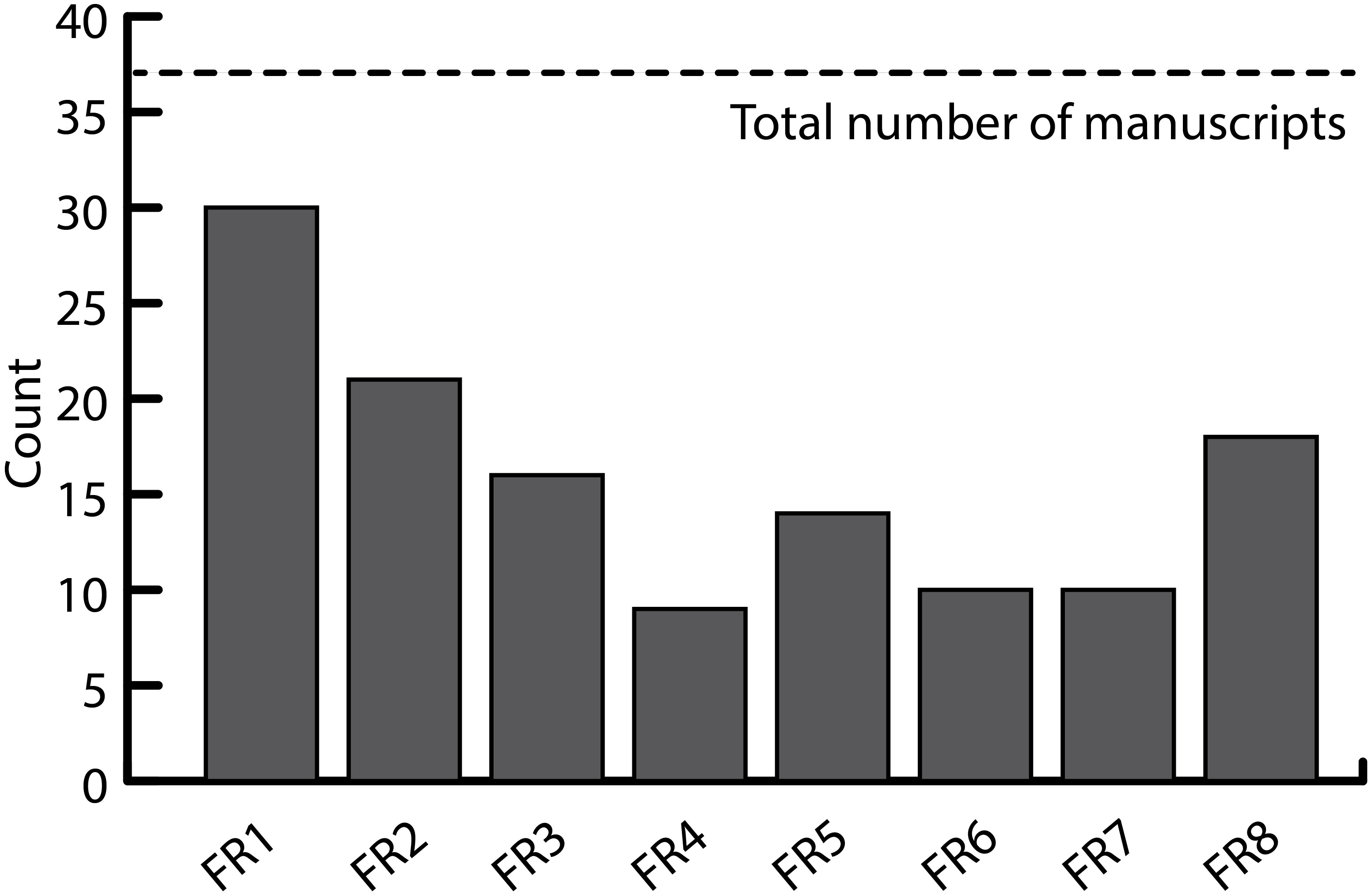}
   \caption{}
   \label{fig:Figure7d} 
\end{subfigure}

\caption{Overview of the (a) selected literature after the three-stage filtering procedure, (b) DT capabilities within the selected manuscripts, (c) prevalence of IRs across the selected manuscripts, and (d) prevalence of FRs across the selected manuscripts.}
\end{figure*}

For the second filtering stage, the criteria provided in Table \ref{Table3} are used to filter the abstracts leading to a total of 140 manuscripts staged for further analysis across Springer (41), ScienceDirect (22), ACM (19), IEEE Xplore (27), and others (31). For the thrid filtering stage and using the criteria provided in Table \ref{Table4}, a total of 37 manuscripts were earmarked across Springer (5), ScienceDirect (13), ACM (1), IEEE Xplore (6), and others (12), as shown in Fig.~\ref{fig:Figure7a}. The literature selected for this study primarily covers PMx applications within three sectors: aviation and automobiles, energy and utilities, and manufacturing and production.

%%%%%%%%%%%%%%%%%%%%%%%
%%%%% SECTION 5 B %%%%%
%%%%%%%%%%%%%%%%%%%%%%%

\subsection{Prevalence of Requirements and Associated Challenges}
\label{sec:Section5-B}
In this section, we analyze the 37 selected manuscripts to determine how each paper addresses the requirements identified in Section~\ref{sec:Section4}. Each manuscript is carefully studied to determine the extent to which it fulfills the specified requirements. Upon completing this comprehensive review, we aggregate the findings for each requirement, highlighting the number of manuscripts that satisfy each criterion. This evaluation serves as an initial indicator of gaps in the current body of literature, emphasizing the requirements that are less addressed or studied, as well as in which contexts. The results also aim to substantiate the legitimacy of the capability grid proposed in Fig.~\ref{fig:Figure5}. We hope that this analysis will provide a roadmap for future research, pointing to the areas that warrant further investigation to support the progress and maturation of PMx DTs.

The distributions of IRs are shown in Fig.~\ref{fig:Figure7c}. Among the selected literature, the most frequently encountered IRs are IR1 and IR4. IR1 pertains to the physical properties of the asset -- such as geometry and material properties -- and stands as an essential component. Creating a virtual representation of the physical system necessitates the inclusion of these fundamental physical characteristics. Similarly, IR4, which describes performance metrics, is a prevalent requirement. Accurately capturing the status of the physical asset is crucial, regardless of the task at hand. These metrics not only serve as critical indicators of the asset's health, condition, and performance, but also foster a shared understanding of the current status of the asset among all stakeholders. 

Although IR3, IR5, and IR6 are not utilized as frequently, they play a crucial (yet sometimes overlooked) role in shaping a more comprehensive and effective PMx strategy. IR3 pertains to contextual information and is particularly valuable in situations where the asset operates under various circumstances, allowing for more situation-aware decision making. However, practically including all relevant context within the DT is often unfeasible. IR5 relates to the availability of the asset's historical data, enhancing the resilience of PMx systems by providing empirical validation of maintenance strategies and delineating system performance trends. Nonetheless, there may be instances where historical data is not readily accessible. In such cases, DTs can be constructed based on a purely theoretical foundation, simulating the physical and operational behavior of the system in a simulated setting. 

IR6 focuses on the identification and understanding of system faults, playing a pivotal role in ensuring alignment among diverse stakeholders involved in PMx DTs. In complex systems where faults can manifest in various forms and new types of faults may emerge, explicitly defining all potential faults can be challenging and might limit the system's versatility. However, neglecting to define these faults could increase the likelihood of miscommunication among stakeholders. Therefore, establishing a clear understanding of the specific faults that the system aims to detect is crucial for maintaining a unified approach and preventing confusion in operational settings. 

The most frequently overlooked IR is IR2, which relates to the structural or material references of the asset. As previously discussed in Section~\ref{sec:Section4}, this may be due to a combination of factors, such as the absence of high-fidelity reference values (e.g., run-to-failure data or lifetime data), along with challenges in establishing high-fidelity thresholds. As IR2 necessitates that the reference values be dynamic, there is a lack of understanding in how to make these values within the DT dynamically updatable.

The distributions of FRs are shown in Fig.~\ref{fig:Figure7d}. The FR most frequently observed is FR1, which concerns the capability of the DT to maintain a consistent representation of the physical system. This characteristic is pivotal in ensuring that the DT reliably mirrors the behavior of its real-world counterpart across various operating conditions. Manuscripts lacking emphasis on FR1 often rely heavily on incoming data, potentially limiting practical application. In these cases, the DT might struggle to provide accurate results when faced with data beyond its typical operational range. 

FR2, FR3, FR5, and FR8, though less frequently addressed than FR1, have garnered considerable attention in the selected literature. FR2 involves the system's adaptability to diverse operational scenarios. This aspect is often overlooked due to insufficient data about the asset's operational and environmental conditions, limiting the system's ability to comprehend and adapt to operational contexts, including unanticipated changes in production demand or power surges, which are integral parts of operational contexts in shop floor maintenance. 

FR3 pertains to the interpretability of the system's output. Systems heavily reliant on complex ML or AI models, such as deep
learning, might pose challenges due to the black-box nature of these models, hindering interpretability. 
Currently, when creating PMx DTs, interpretable models are often employed as an initial approach to address the poor explainability of data-driven methods in PMx~\cite{kapteyn2020toward,tripura2022probabilistic}. The integration of model-agnostic add-ons represents an emerging approach, offering a more flexible solution to enhance the explainability of advanced systems. This trend highlights the growing recognition of the importance of FR3, emphasizing the need for transparent and understandable decision-making processes in PMx DT systems.

FR5 focuses on the system's ability to adapt to evolving circumstances throughout an asset's lifecycle. Challenges arise from the lack of comprehensive and relevant data, which hinders the system's ability to adapt to concept drift. Additionally, the pursuit of model simplicity often results in the use of static models, which fail to account for concept drift.

FR8 involves a system's ability to recognize and quantify inherent uncertainties, which are often underrepresented in complex systems. This underrepresentation can stem from an incomplete understanding of uncertainty or a lack of tools for quantification, such as unknown causal relationships between variables or the multi-modality of behaviors.

Most reviewed literature overlooks FR4, FR6, and FR7. FR4 refers to the system's capacity to function amidst potential disturbances, where consideration for errors like DA errors, input/output errors, and internal system errors is lacking. There is a perception that integrating these considerations into PMx DT operations might diminish the systems's autonomy, but discussions around this topic are imperative as human intervention is not yet fully excluded. FR6 refers to a system's efficiency across various scales. Some studies focus on case studies primarily as proof-of-concept. These studies conduct research on a smaller scale based solely on data volume that is convenient for experimental purposes, thereby not fully accounting for the diverse, multivariate environments characteristic of real-world applications. FR7 concerns a system's ability to uphold its functionalities across more diverse assets than initially trained for. This aspect is often neglected in favor of validating a concept or probing specific phenomena, ignoring transferability to secondary roles.

\begin{figure*}[t]
\centering
\begin{subfigure}[b]{0.34\textwidth}
   \includegraphics[width=1\linewidth]{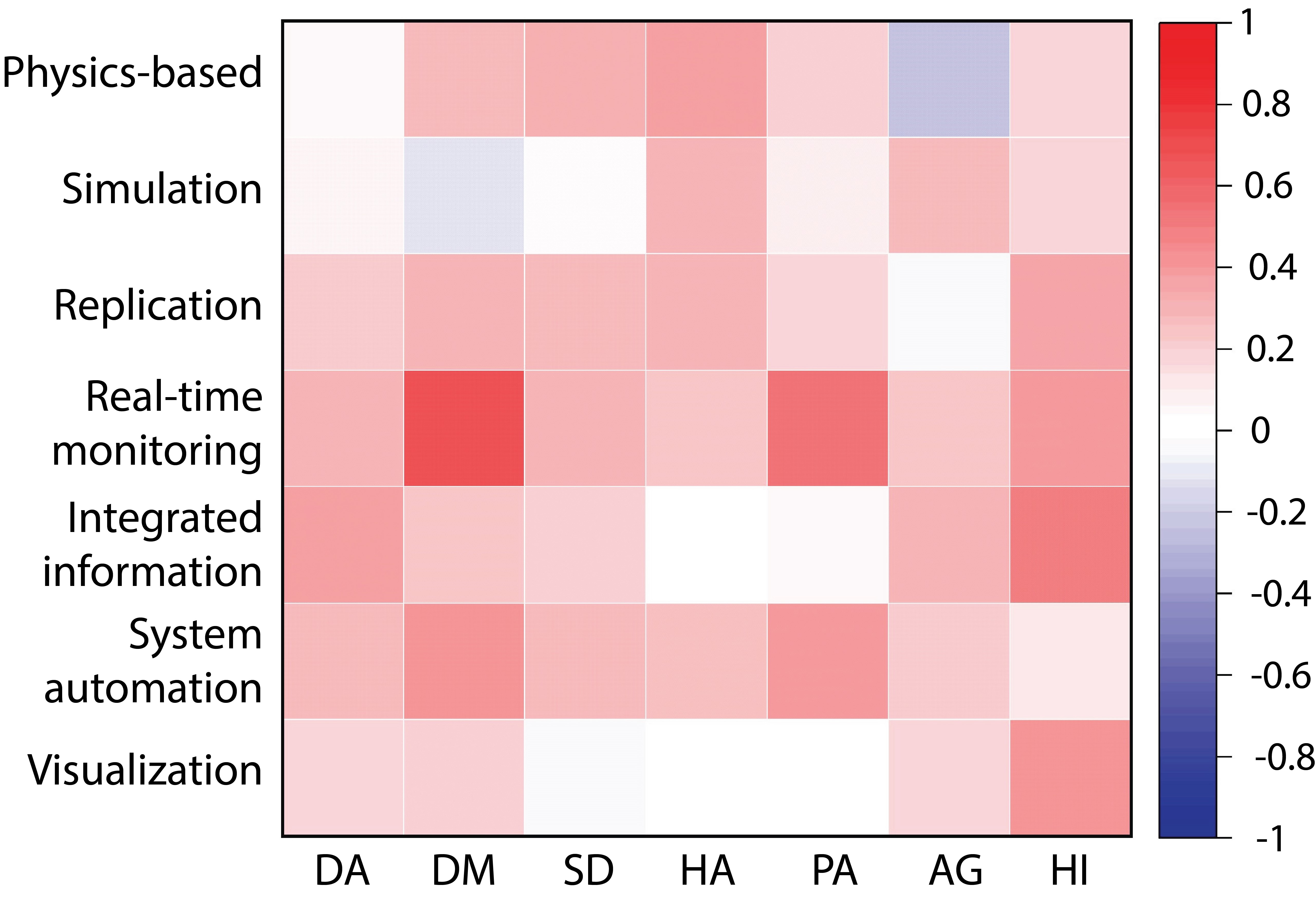}
   \caption{}
   \label{fig:Figure8a} 
\end{subfigure}
\hfill
\begin{subfigure}[b]{0.34\textwidth}
   \includegraphics[width=1\linewidth]{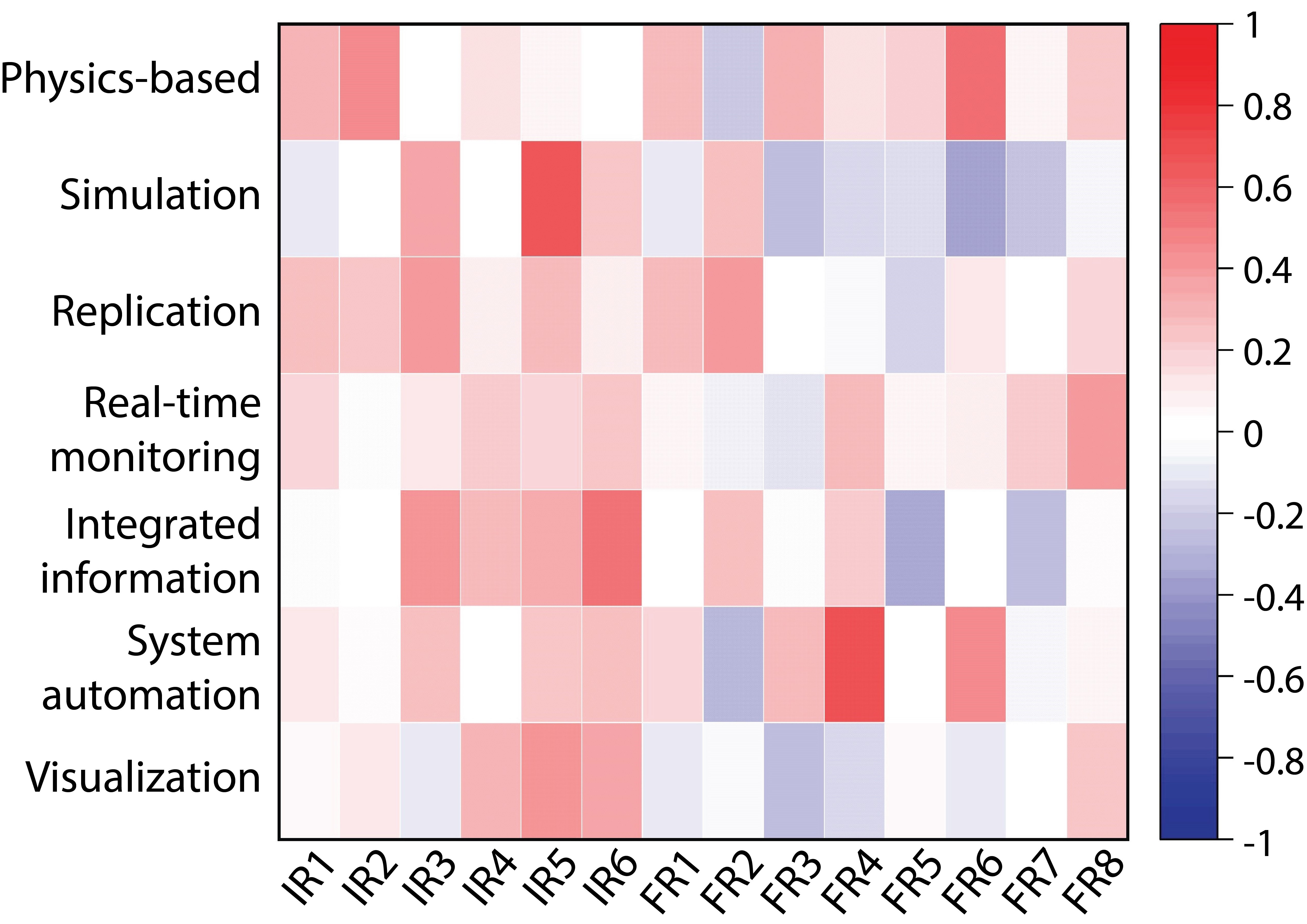}
   \caption{}
   \label{fig:Figure8b}
\end{subfigure}
\hfill
\begin{subfigure}[b]{0.29\textwidth}
   \includegraphics[width=1\linewidth]{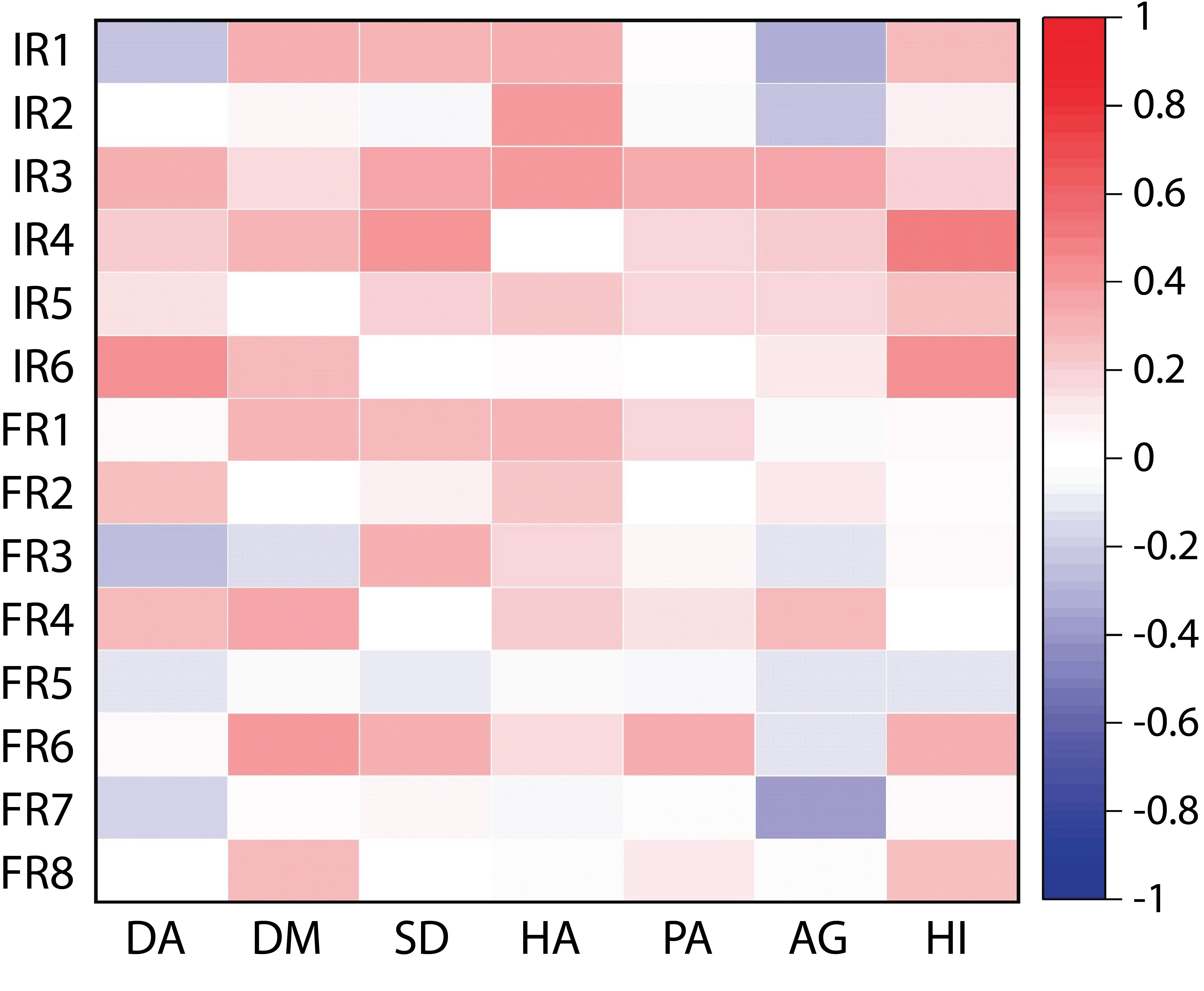}
   \caption{}
   \label{fig:Figure8c}
\end{subfigure}

\caption{Pearson correlation coefficient between (a) PMx modules and DT capabilities, (b) requirements and DT capabilities, and (c) PMx modules and requirements.}
\end{figure*}

Next, we present the Pearson correlation coefficients derived from the selected literature to illustrate the relationships between the PMx modules and the DT capabilities (Fig.~\ref{fig:Figure8a}), the requirements and the DT capabilities (Fig.~\ref{fig:Figure8b}), and the requirements and the PMx modules (Fig.~\ref{fig:Figure8c}). The similarities between Fig.~\ref{fig:Figure8a} and the capability grid presented in Fig.~\ref{fig:Figure5} substantiate the legitimacy of the proposed capability grid that grounded the IR and FR selection methodology. Fig.~\ref{fig:Figure8b} and Fig.~\ref{fig:Figure8c} represent a snapshot of the current state of PMx DT technology, rather than illustrating the ideal connections between the requirements and standardized efforts; a comparison between the current and ideal states underscores the key research gaps. Table~\ref{Table5} offers a more comprehensive view of how these requirements should be associated with the respective DT capabilities or PMx modules. In this table, requirements that are not bolded denote those requirements that have been investigated and met, to some extent, when the corresponding capabilities are implemented as part of a PMx DT. This is indicated by a Pearson coefficient of $\geq0.25$. Conversely, requirements set in bold text represent areas that demand further study, as they have not been satisfactorily addressed in the context of their associated capabilities.

\begin{table}[h]
\caption{\textbf{Relationships between requirements, DT capabilities, and PMx modules.}}
\setlength{\tabcolsep}{3pt}
\centering
\begin{tabular}{|p{60pt}|p{180pt}|}
\hline
DT capability&
Requirements necessary to fulfill DT capability\\
\hline
    Physics-based & IR1, IR2, \textbf{IR3}, IR6, FR1, FR3, \textbf{FR4}, FR5, FR6\\
    Simulation & IR3, IR5, IR6, \textbf{FR2}, \textbf{FR8}\\
    Replication & IR1, IR2, IR3, IR5, FR1, FR2\\
    Real-time monitoring & \textbf{IR3}, IR4, IR5, IR6, FR4, \textbf{FR6}, FR8\\
    Integrated information & \textbf{IR1},\textbf{IR2},IR3, IR4, IR5, IR6\\
    System automation & FR3, FR4, FR6\\
    Visualization & IR4, IR5, IR6, \textbf{FR3}, \textbf{FR8}\\
    \hline
PMx module & Requirements necessary to fulfill PMx module\\
    \hline
    DA & \textbf{IR2}, IR3, IR6, \textbf{FR1}, FR2, FR4, \textbf{FR5}\\
    DM & IR1, IR4, \textbf{IR5}, IR6, FR1, FR4, FR6, FR8\\
    SD & IR1, \textbf{IR2}, IR3, IR4, IR5, FR1, \textbf{FR2}, FR3, FR4, FR5, FR6, FR7, FR8\\
    HA & IR1, IR2, IR3, \textbf{IR4}, IR5, \textbf{IR6}, FR1, FR2, FR3, FR4, \textbf{FR5}, FR6, \textbf{FR7}, \textbf{FR8}\\
    PA & \textbf{IR1}, \textbf{IR2}, IR3, IR4, IR5, \textbf{IR6}, FR1, \textbf{FR2}, \textbf{FR3}, FR4, \textbf{FR5}, FR6, \textbf{FR7}, \textbf{FR8}\\
    AG & IR3, IR4, IR5, \textbf{FR3}, FR4, \textbf{FR7}\\
    HI & FR4, FR6, \textbf{FR3}, FR6, FR8\\
\hline
\end{tabular}
\label{Table5}
\end{table}

%%%%%%%%%%%%%%%%%%%%%
%%%%% SECTION 6 %%%%%
%%%%%%%%%%%%%%%%%%%%%

\section{Conclusion}
\label{sec:Section6}

This paper sets the foundation for PMx DT and paves the way for future investigations by systematically identifying existing gaps between the informational and functional requirements of PMx systems and the capabilities of DTs. Addressing these gaps is essential for transitioning PMx DTs from a novel, research-focused field into a mature, industry-standard solution.

Through our unique capability grid, we have established a clear connection between PMx tasks and DT technologies, aiding in a better understanding of how DTs can support various PMx processes. Additionally, our curated taxonomy of PMx DTs provides valuable insights into the challenges encountered when designing these systems. Our structured approach has identified several critical requirements that have been overlooked in the existing PMx DT literature. For example, IR1 and IR2, which focus on physical properties and reference values, need further exploration to enhance integrated information capabilities. Similarly, FR3 and FR8, dealing with interpretability and uncertainty, require more extensive research to advance visualization and simulation capabilities. Beyond individual modules, the broader challenge lies in ensuring seamless interaction and integration within the holistic PMx DT system. This challenge includes addressing the inherent technological complexity of DTs, concerns about data quality and integrity, and issues related to scalability and sample inefficiency. Moreover, security concerns, the need for improved interoperability between systems, and a significant gap in the necessary skill sets further complicate the full realization of DTs in the PMx domain.

To summarize our discussion on PMx and DTs, it is crucial to reflect on the foundational successes of standardized roadmaps, as highlighted in  Section~\ref{sec:Section1}. The software industry's achievement in establishing standard IRs and FRs, which guided the complexities of development, serves as a significant precedent. This precedent mirrors our efforts in the realm of PMx and DTs, emphasizing the symbiotic relationship between theory and application and demonstrating how frameworks can effectively address major challenges. 

By identifying how current DT technologies meet the IRs and FRs of PMx systems, our analysis not only highlights the practical applications of DTs in PMx but also illustrates the intricate integration between DT technology and PMx strategies. By detailing the specific IRs and FRs that state-of-the-art PMx DT systems are equipped to fulfill, we provide structured insights into how DTs can enhance maintenance processes. This alignment of DT capabilities with PMx needs reflects a deep integration of theoretical principles with practical maintenance strategies, forming a cornerstone of advanced maintenance management.

For a more detailed exploration of the models, software, and components involved in this integration, as well as their interactions within the PMx framework, readers are encouraged to consult our previous work, which delves into the specifics of these integrations and their implications for the development of robust PMx systems~\cite{ma2024stateoftheart}. To fully harness the potential of DTs for PMx automation, we recommend the following practical steps for standardization based on the findings:
\begin{enumerate}
    \item Capabilities as benchmarks: Establish industry benchmarks based on the distinct DT capabilities we have associated with PMx. These benchmarks will guide the development and implementation of DTs customized for PMx automation tasks, illustrating the application of theoretical concepts in setting practical standards that reflect the detailed capabilities of DTs in addressing PMx needs.
    \item Guideline framework from IRs and FRs: Develop a standardized framework or protocol based on the identified IRs and FRs to ensure that DTs are designed and operated in alignment with the requirements of PMx systems. This framework will serve as a practical example of integrating DT technology with PMx strategies, demonstrating how theoretical models are translated into operational guidelines.
    \item Standardized mapping for integration: Advocate for the adoption of a mapping methodology to ensure consistent integrations of PMx standards and DT capabilities. This facilitates a seamless and efficient automation process.
    \begin{itemize}
        \item Expanding upon previous preliminary works that delineates the levels of autonomy within the PMx
domain -- inspired by the tiered autonomy models in the autonomous vehicle field -- we propose a strategic linkage between IRs,
FRs, and DT capabilities tailored to the specific autonomy level of a PMx system~\cite{flanigan2022autonomy}. This proposed
standardization outlines, for each level of autonomy, the necessary IRs and FRs that must be met to enable the corresponding
DT capabilities. Such standardization efforts offer stakeholders a clearer path at this pioneering yet complex crossroad
between PMx and DTs, aiming to align expectations in the face of the rapidly evolving landscape of intelligent maintenance
systems.
    \end{itemize}
    \item Collaborative feedback mechanisms: Establish platforms that facilitate the continuous exchange of insights among academia, industry, and technology developers. Similar to the efforts of organizations like Building Smart International in the smart building industry, these collaborative platforms have proven essential in driving the digital transformation of various sectors. This ensures that automation standards and practices advance concurrently with the rapidly progressing technological landscape. By fostering a feedback loop, this mechanism underscores the practical application of theoretical research, facilitating continuous improvement and adaptation of DT integration strategies in real-world settings.
\end{enumerate}

Building upon these standardization guidelines, we propose a forward-looking research agenda. The vision and trajectory outlined in this paper are not only prescriptive for the wider field but also embody our own research path. This agenda reinforces the deep integration between DT technologies and PMx strategies, underlining how theoretical advancements can lead to practical applications in maintenance management. The foundational understanding provided in this paper not only identifies existing gaps but also prepares the stage for impactful, real-world case studies in our subsequent investigations. Future research should primarily focus on the neglected IRs and FRs, particularly those crucial for advancing specific DT capabilities. Collaborative efforts across disciplines, including data science, ML, and systems engineering, can yield novel insights and methods to meet these requirements. Furthermore, a greater emphasis should be placed on the systemic view of PMx DTs. Studies should not only address requirements within a specific module but also explore the interrelationships between these modules and how improvements in one may affect the others. Lastly, as DTs evolve, the intersection of DT characteristics with PMx requirements presents a fertile ground for research, highlighting the ongoing need to refine and expand the theoretical models that guide DT implementations in PMx contexts as explored in detail in our prior work \cite{ma2024stateoftheart}. Consider the following DT characteristics as examples:
\begin{itemize}
    \item Interoperability: Future studies should address the development of standardized protocols that allow for seamless data exchange on IRs between different DT systems and external platforms. This is crucial for integrating PMx data across various stages of assets' lifecycle.
    \item Fidelity: Future studies should investigate advanced simulation technologies that improve the accuracy and detail of DT models. Research should focus on enhancing the resolution through which DTs replicate physical systems, which could be solved by tackling requirements necessary for Simulation and Replication mentioned in Table~\ref{Table5}.
    \item Scalability: Future studies should explore methods to efficiently scale DT implementations across different machine types and environments within an industry, ensuring that solutions are adaptable to both small-scale operations and large industrial complexes. This DT characteristic could be solved by tackling requirements necessary for System Automation mentioned in Table~\ref{Table5}.
\end{itemize}
Addressing the identified gaps mentioned in this section is expected to shift PMx DTs from academic novelty to practical industry application, ultimately enhancing the effectiveness and efficiency of PMx systems. 

\section*{CRediT Authorship Contribution Statement}
\noindent \textbf{Sizhe Ma:} Conceptualization, Methodology, Software, Validation, Formal Analysis, Investigation, Data Curation, Writing - Original Draft.
\textbf{Katherine A. Flanigan:} Conceptualization, Methodology, Resources, Writing - Original Draft, Writing - Review \& Editing, Visualization, Supervision, Project Administration, Funding Acquisition.
\textbf{Mario Berg\'es:} Conceptualization, Methodology, Resources, Writing - Review \& Editing, Supervision, Project Administration, Funding Acquisition.

\section*{Funding}
 \noindent Information developed under this Award was or is sponsored by the U.S. Army Contracting Command under Contract Number W911NF20D0002 and W911NF22F0014 delivery order No. 4.

\section*{Declaration of Competing Interest}
 \noindent The authors declare no conflict of interest.

%%       \cite{<label>} ==> [1]

%%
%%Example citation, See \cite{lamport94}.

\bibliographystyle{elsarticle-num} 
\bibliography{reference}



\end{document}